# General Fourier Feature Physics-Informed Extreme Learning Machine (GFF-PIELM) for High-Frequency PDEs

Dr Fei **Ren**
Email: ren87@outlook.com
a. School of Mechanical Engineering, Shandong Key Laboratory of CNC Machine Tool Functional Components，Qilu University of Technology (Shandong Academy of Sciences), Jinan 250353, China；　b. Shandong Institute of Mechanical Design and Research, Jinan 250031, China

Dr Sifan **Wang**
Email: sifan.wang@yale.edu
Institution for Foundation of Data Science, Yale University, New Haven, CT 06520

Professor Pei-Zhi **Zhuang**
E-mail: zhuangpeizhi@sdu.edu.cn
School of Qilu Transportation, Shandong University, Jinan, 250002, China

Prof Hai-Sui **Yu**, FREng
E-mail: Yu.H@leeds.ac.uk
School of Civil Engineering, University of Leeds, Leeds, LS2 9JT, UK

Dr He **Yang**
Email: yanghesdu@mail.sdu.edu.cn
Corresponding author
School of Qilu Transportation, Shandong University, Jinan, 250002, China

**Abstract:**

Conventional physics-informed extreme learning machine (PIELM) often faces challenges in solving partial differential equations (PDEs) involving high-frequency and variable-frequency behaviors. To address these challenges, we propose a general Fourier feature physics-informed extreme learning machine (GFF-PIELM). We demonstrate that directly concatenating multiple Fourier feature mappings (FFMs) and an extreme learning machine (ELM) network makes it difficult to determine frequency-related hyperparameters. Fortunately, we find an alternative to establish the GFF-PIELM in three main steps. First, we integrate a variation of FFM into ELM as the Fourier-based activation function, so there is still one hidden layer in the GFF-PIELM framework. Second, we assign a set of frequency coefficients to the hidden neurons, which enables ELM network to capture diverse frequency components of target solutions. Finally, we develop an innovative, straightforward initialization method for these hyperparameters by monitoring the distribution of ELM output weights. GFF-PIELM not only retains the high accuracy, efficiency, and simplicity of the PIELM framework but also inherits the ability of FFMs to effectively handle high-frequency problems. We carry out five case studies with a total of ten numerical examples to highlight the feasibility and validity of the proposed GFF-PIELM, involving high frequency, variable frequency, multi-scale behaviour, irregular boundary and inverse problems. Compared to conventional PIELM, the GFF-PIELM approach significantly improves predictive accuracy without additional cost in training time and architecture complexity. Our results confirm that that PIELM can be extended to solve high-frequency and variable-frequency PDEs with high accuracy, and our initialization strategy may further inspire advances in other physics-informed machine learning (PIML) frameworks.

# 1. Introduction

The rapid advancement of artificial intelligence and data science has substantially accelerated the development of physics-informed machine learning (PIML) in recent years. This emerging method can help solve ordinary/partial differential equations (ODEs/PDEs) with or without measured data by leveraging the universal approximation capability of machine learning (Karniadakis et al. 2021). Compared to conventional numerical methods, PIML collocates training points without mesh generation, thus allowing it to address challenges such as mesh detorsion and complex solution domains. Moreover, its seamless integration of data and physics also provides notable advantages in solving inverse problems. These advantages make PIML a research frontier across computational science, solid and fluid mechanics, biology, geology, electromagnetics, and atmospheric sciences, among others (Cai et al. 2021; Cuomo et al. 2022; Yuan et al. 2025).

One of the most prominent branches of PIML is the physics-informed neural networks (PINNs) proposed by Raissi et al. (2019), and many case studies have demonstrated the power of PINNs for solving various ODEs/PDEs (Mao et al. 2020; Cai et al. 2021; Chen et al. 2021; Song et al. 2024). In spite of the remarkable success, the limitations of PINNs cannot detract from the overall excellence. Very often PINNs cannot provide acceptably accurate solutions or even fail to train especially for those PDEs exhibiting high-frequency or multi-scale behavior (Wang et al. 2022; Sallam and Fürth 2023; Song and Wang 2023), and their training efficiency is generally low (Cuomo et al. 2022).

In terms of the first-mentioned limitation of PINNs, the weakness of neural networks in learning multi-frequency or high-frequency functions is referred to as the spectral bias, which means that neural networks tend to capture low-frequency components of input data (Cao et al. 2019; Rahaman et al. 2019; Xu et al. 2019; Geifman et al. 2022). Using neural tangent kernel (NTK) theory (Jacot et al. 2018), Wang et al. (2022) demonstrated that spectral bias is indeed a prevalent issue in PINNs and constitutes a primary obstacle to the accurate approximation of high-frequency and multi-scale functions. Currently there are two main strategies developed to alleviate this problem:

(i) Add a Fourier feature mapping (FFM) layer as the first hidden layer of the network. Tancik et al. (2020) incorporated a simple FFM into fully connected networks. They showed that tuning Fourier feature parameters to suitable values enables the network to better capture high-frequency components and effectively alleviate spectral bias. Subsequently, FFM has also been introduced to PINNs for solving PDEs with high-frequency or multi-scale behavior (Wang et al. 2021; Jin et al. 2024; Li et al. 2024; Zhang et al. 2024).

(ii) Adjust the frequency of neural networks via modifying activation functions. Sitzmann et al. (2020) employed sinusoidal activations to introduce periodicity into deep neural networks. Based on this idea, Zhang (2023) proposed the triangular deep neural network with activation function $\cos(\lambda x)$, where choosing a proper $\lambda$ raises the frequency of neural networks.

It is interesting to find that both strategies share the common philosophy that we should adjust the frequency of neural networks to match the frequency of target functions. However, these methods inevitably introduce additional hyperparameters related to the intrinsic frequency of networks, and we term them frequency coefficients. We must identify the frequency that networks prefer to learn by tuning these hyperparameters. In practice the intrinsic PDE frequency is usually unknown in forward problems, and selecting additional hyperparameters can be computationally expensive and largely reliant on trial-and-error procedures (e.g., grid searching method). Besides, these strategies may face challenges when solving PDEs with variable frequency (e.g., from low to high frequency), as it becomes more difficult to "guess" the frequency coefficients across wider frequency ranges.

Regarding the second-mentioned limitation of PINNs, the deep neural network architecture and the time-consuming gradient-descent method could be two important reasons for the low training efficiency of PINNs. An alternative is to replace deep neural networks with the extreme learning machine (ELM) (Huang et al. 2006), leading to the physics-informed extreme learning machine (PIELM) (Dwivedi and Srinivasan 2020). Increasing publications have proven that PIELM can not only drastically reduce training time but also improve solution accuracy (Calabrò et al. 2021; Dong and Li 2021; Schiassi et al. 2021; Liu et al. 2023; Dwivedi et al. 2025; Mishra et al. 2025; Peng et al. 2025; Ren et al. 2025; Rout 2025; Wang et al. 2025; Zhu et al. 2025). For instance, Schiassi et al. (2021) introduced a hard-constrained PIELM method by integrating the theory of functional connections and PIELM, and it outperforms PINNs in accuracy by 4 to 11 orders of magnitude for linear and nonlinear bi-dimensional PDEs. Ren et al. (2025) solved the nonlinear Stefan problems by iterative PIELM, which saves more than 98% of training time and improves accuracy by a factor of $10^2 \sim 10^4$. Nevertheless, difficulty arises when we utilize PIELM to solve PDEs with high-frequency and variable-frequency behavior. The input weights in PIELM are randomly generated and fixed in network training, and the frequency of the ELM network is predominantly determined once the input weights are assigned. In other words, we must carefully select appropriate frequency coefficients for initialization, making the direct concatenation of FFM within ELM a suboptimal choice.

Concerning the limitations of PINN and PIELM for solving high-frequency and variable-frequency PDE problems, we propose a general Fourier feature PIELM (GFF-PIELM) approach. The primary contributions of this paper are summarized as follows:

(i) We integrate an FFM variant into the ELM network, by employing the Fourier-based activation function and a set of frequency coefficients. This method can effectively solve high-frequency and variable-frequency PDEs while preserving the simplicity of PIELM.

(ii) We propose a dedicated initialization method for selecting frequency coefficients. This method can capture the initially unknown frequency components of target functions, thereby avoiding the conventional trial-and-error process.

(iii) We conduct extensive experiments on challenging PDE problems involving high frequency, variable frequency, irregular boundaries, inverse problems, and multi-scale behavior. The results demonstrate that the proposed GFF-PIELM achieves substantial improvements over standard PIELM.

The rest of the paper is organized as follows. Section 2 presents some preliminaries, including the PIELM method and FFM technique. In Section 3, we first outline our proposed GFF-PIELM framework, and then introduce an initialization method for choosing frequency-related hyperparameters. In Section 4, we perform a series of case studies to validate the performance of GFF-PIELM. Section 5 discusses its limitations. Finally, Section 6 summarizes the main conclusions.

## 2. Preliminaries for PIELM and Fourier feature mapping

This section presents a brief overview of physics-informed extreme learning machine (PIELM) and the FFM method in machine learning.

### *2.1. Physics-informed extreme learning machine*

Let us consider a general form of PDEs as follows:

$$\mathcal{D}\left[u(\boldsymbol{x},t)\right] = f(\boldsymbol{x},t), \ \boldsymbol{x} \in \Omega \text{ and } t \in [0,T] \tag{1}$$

$$\mathcal{I}\left[u(\boldsymbol{x},0)\right] = u_0(\boldsymbol{x}), \ \boldsymbol{x} \in \Omega \tag{2}$$

$$\mathcal{B}\left[u(\boldsymbol{x},t)\right] = g(\boldsymbol{x},t), \ \boldsymbol{x} \in \partial\Omega \text{ and } t \in [0,T] \tag{3}$$

where $\boldsymbol{x} \in \mathbb{R}^d$ and $t \in [0,T]$ denote space and time coordinates; $\Omega$ denotes the computational domain in $\mathbb{R}^d$ with boundary $\partial\Omega$; $\mathcal{D}$ stands for the partial differential operator, $\mathcal{I}$ and $\mathcal{B}$ are the initial and boundary condition operators; $f(\boldsymbol{x}, t)$, $u_0(\boldsymbol{x})$ and $g(\boldsymbol{x}, t)$ are the source term, initial condition and boundary condition, respectively.

As a fast version of PINN, PIELM uses ELM instead of deep neural networks to approximate the latent solution (Dwivedi and Srinivasan 2020). ELM is a single-layer feed-forward neural network which randomly assigns input layer weights and analytically determines the output layer weights. The PIELM is schematically shown in Figure 1. The latent solution $u(\boldsymbol{x}, t)$ can be expressed as follows:

$$u(\boldsymbol{x},t;\boldsymbol{\beta}) = \sum_{m=1}^{M} \beta_m \phi\left(\boldsymbol{w}_m^{\mathrm{T}}[\boldsymbol{x},t]^{\mathrm{T}} + b_m\right) = \left[\phi\left(\boldsymbol{W}^{\mathrm{T}}[\boldsymbol{x},t]^{\mathrm{T}} + \boldsymbol{b}\right)\right]^{\mathrm{T}} \boldsymbol{\beta} \tag{4}$$

where $M$ is the number of neurons in the hidden layer; $\phi$ denotes the activation function; $\boldsymbol{W} = [\boldsymbol{w}_1, \boldsymbol{w}_2, ..., \boldsymbol{w}_M] \in \mathbb{R}^{(d+1) \times M}$ is the input layer weight matrix where $\boldsymbol{w}_m = [w_{m,1}, w_{m,2}, ..., w_{m,d+1}]^{\mathrm{T}}$ for $m = 1, 2, ..., M$; $\boldsymbol{b} = [b_1, b_2, ..., b_M]^{\mathrm{T}}$ is the bias vector; $\boldsymbol{\beta} = [\beta_1, \beta_2, ..., \beta_M]^{\mathrm{T}}$ is the output layer weight vector. Generally, $\boldsymbol{W}$ and $\boldsymbol{b}$ are uniformly initialized within an interval $[-L, L]$ and remain fixed throughout training, and $\boldsymbol{\beta}$ is trainable and determined by the least squares method with the Moore-Penrose generalized inverse. The output of $m$-th hidden neuron can be represented by $h_m(\boldsymbol{x},t) = \phi\left(\boldsymbol{w}_m^{\mathrm{T}}[\boldsymbol{x},t]^{\mathrm{T}} + b_m\right)$. Then, the partial derivatives of $u$ with respect to independent variables are expressed as follows:

$$\begin{cases} \dfrac{\partial^n u}{\partial x_i^n} = \sum_{m=1}^{M} \beta_m \dfrac{\partial^n h_m}{\partial x_i^n} = \sum_{m=1}^{M} \beta_m w_{m,i}^n \dfrac{\partial^n \phi}{\partial \left(\boldsymbol{w}_m^{\mathrm{T}}[\boldsymbol{x},t]^{\mathrm{T}} + b_m\right)^n}, \quad i = 1, 2, ..., d \\ \dfrac{\partial^n u}{\partial t^n} = \sum_{m=1}^{M} \beta_m \dfrac{\partial^n h_m}{\partial t^n} = \sum_{m=1}^{M} \beta_m w_{m,d+1}^n \dfrac{\partial^n \phi}{\partial \left(\boldsymbol{w}_m^{\mathrm{T}}[\boldsymbol{x},t]^{\mathrm{T}} + b_m\right)^n} \end{cases} \tag{5}$$

where $n$ refers to the $n$-th order derivative.

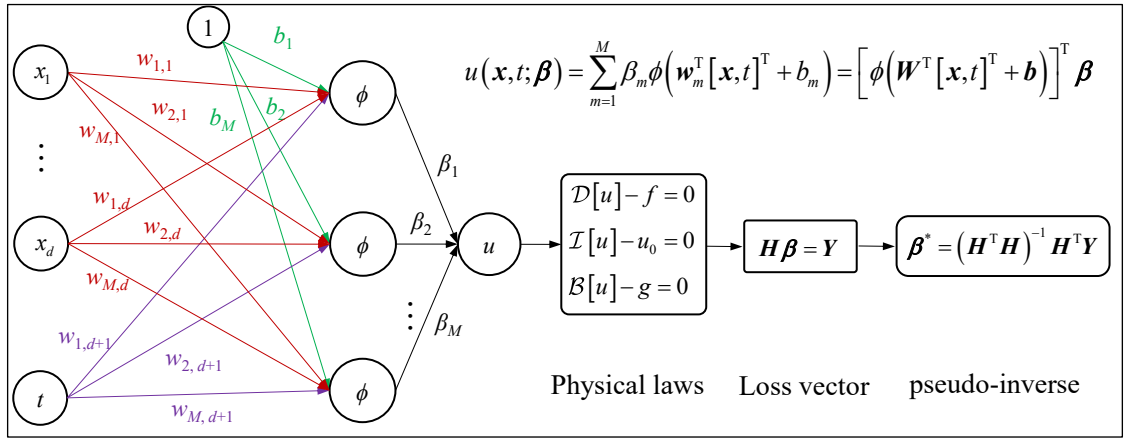

**Figure 1 Schematic diagram of PIELM**

In particular, if $\mathcal{D}$, $\mathcal{I}$ and $\mathcal{B}$ in Eqs. (1), (2) and (3) are linear differential operators (i.e., linear problems), the PDEs, initial conditions and boundary conditions yield a system of linear equations in a matrix form as $\boldsymbol{H\beta} = \boldsymbol{Y}$, where

$$\boldsymbol{H} = \begin{bmatrix} \mathcal{D}[h_1(\boldsymbol{x}_{C,1},t_{C,1})] & \cdots & \mathcal{D}[h_M(\boldsymbol{x}_{C,1},t_{C,1})] \\ \vdots & \cdots & \vdots \\ \mathcal{D}[h_1(\boldsymbol{x}_{C,N_C},t_{C,N_C})] & \cdots & \mathcal{D}[h_M(\boldsymbol{x}_{C,N_C},t_{C,N_C})] \\ \mathcal{B}[h_1(\boldsymbol{x}_{B,1},t_{B,1})] & \cdots & \mathcal{B}[h_M(\boldsymbol{x}_{B,1},t_{B,1})] \\ \vdots & \cdots & \vdots \\ \mathcal{B}[h_1(\boldsymbol{x}_{B,N_B},t_{B,N_B})] & \cdots & \mathcal{B}[h_M(\boldsymbol{x}_{B,N_B},t_{B,N_B})] \\ \mathcal{I}[h_1(\boldsymbol{x}_{I,1},0)] & \cdots & \mathcal{I}[h_M(\boldsymbol{x}_{I,1},0)] \\ \vdots & \cdots & \vdots \\ \mathcal{I}[h_1(\boldsymbol{x}_{I,N_I},0)] & \cdots & \mathcal{I}[h_M(\boldsymbol{x}_{I,N_I},0)] \end{bmatrix}_{(N_C+N_B+N_I)\times M} \quad (6)$$

$$\boldsymbol{Y} = \begin{bmatrix} f(\boldsymbol{x}_{C,1},t_{C,1}) \\ \vdots \\ f(\boldsymbol{x}_{C,N_C},t_{C,N_C}) \\ g(\boldsymbol{x}_{B,1},t_{B,1}) \\ \vdots \\ g(\boldsymbol{x}_{B,N_B},t_{B,N_B}) \\ u_0(\boldsymbol{x}_{I,1}) \\ \vdots \\ u_0(\boldsymbol{x}_{I,N_I}) \end{bmatrix}_{(N_C+N_B+N_I)\times 1} \quad (7)$$

in which $\boldsymbol{H}$ and $\boldsymbol{Y}$ are known matrices which contain physical laws; $\{\boldsymbol{x}_{C,i},t_{C,i}\}_{i=1}^{N_C}$ are collocation training points for PDEs sampled inside the domain $\Omega\times[0,T]$; $\{\boldsymbol{x}_{I,k},0\}_{k=1}^{N_I}$ and $\{\boldsymbol{x}_{B,j},t_{B,j}\}_{j=1}^{N_B}$ are training points for initial and boundary conditions, respectively. $N_C$, $N_B$ and $N_I$ are the number of training points for PDEs, boundary conditions and initial conditions, respectively. The involved partial derivatives of $h_m$ in $\boldsymbol{H}$ can be calculated manually or using computational tools like Symbolic or automatic differentiation (AD) routines. Eventually, the optimal output layer weight vector $\boldsymbol{\beta}^*$ can be directly obtained by using the Moore-Penrose generalized inverse matrix of $\boldsymbol{H}$, as $\boldsymbol{\beta}^* = \left(\boldsymbol{H}^{\mathrm{T}}\boldsymbol{H}\right)^{-1}\boldsymbol{H}^{\mathrm{T}}\boldsymbol{Y}$. It can be found that the time-consuming gradient-descent-based training methods are avoided in PIELM. In this paper we refer to this type of PIELM as the vanilla PIELM to distinguish it from our proposed GFF-PIELM.

## 2.2. Fourier feature mapping (FFM)

FFM is a technique used in machine learning to transform input data into a higher-dimensional space using sinusoidal functions (sine and cosine waves). This approach helps neural networks better capture high-frequency patterns in data that would be difficult to learn from raw inputs alone. FFM was first introduced in the work of Rahimi and Recht (2007), where they used random Fourier features to approximate an arbitrary stationary kernel function to speed up the training of kernel machines. After that, FFM was combined with deep neural networks (Tancik et al. 2020) and PINNs (Wang et al. 2021) to achieve significant performance improvements. It has been demonstrated in many recent works (Jin et al. 2024; Li et al. 2024; Zhang et al. 2024) that using FFMs as the first hidden layer can remarkably mitigate the pathology of spectral bias for neural networks. A schematic diagram of Fourier feature neural network is shown in Figure 2 (a), and the FFM function can be expressed as

$$\gamma(\boldsymbol{v}) = \begin{bmatrix} \cos(\boldsymbol{B}\boldsymbol{v}) \\ \sin(\boldsymbol{B}\boldsymbol{v}) \end{bmatrix} \tag{8}$$

where $\boldsymbol{v} = [\boldsymbol{x}, t]^{\mathrm{T}}$ is the input spatial-temporal coordinate vector. The frequency matrix $\boldsymbol{B} \in \mathbb{R}^{\frac{m_F}{2} \times (d+1)}$ is initialized from the Gaussian distribution $\mathcal{N}(0, \delta^2)$. Each row of $\boldsymbol{B}$ defines a frequency vector for the transformation, and $m_F$ is the number of frequency vectors. $\delta$ is the user-defined frequency coefficient which determines the distribution of the Fourier basis frequencies. Moreover, the frequency matrix $\boldsymbol{B}$ can be either trainable or untrainable. An FFM with untrainable $\boldsymbol{B}$ is customarily called the random FFM that is actually more commonly used since it is simple and scarcely increases the complexity of networks. The underlying mechanisms of the FFM technique have been rigorously analyzed by Wang et al. (2021) within the framework of NTK theory. They confirmed that neural networks prioritize learning the target function components along the eigendirections of the NTK that have larger eigenvalues. For conventional fully connected neural networks, the eigenvalues of the NTK decrease monotonically with increasing frequency of their corresponding eigenfunctions, leading to a much slower convergence rate for high-frequency components. For Fourier feature neural networks, the parameter $\sigma$ actually controls the frequency that the network is biased to learn. Larger values of $\sigma$ increase the probability of sampling large-magnitude $\boldsymbol{B}$, which corresponds to high-frequency eigenfunctions and narrower eigenvalue gaps. Thus, Fourier feature can mitigate spectral bias and accelerate the learning of high-frequency components of the target function. It should be emphasized that excessively large values of $\sigma$ may cause over-fitting during initialization of the Fourier feature mapping.

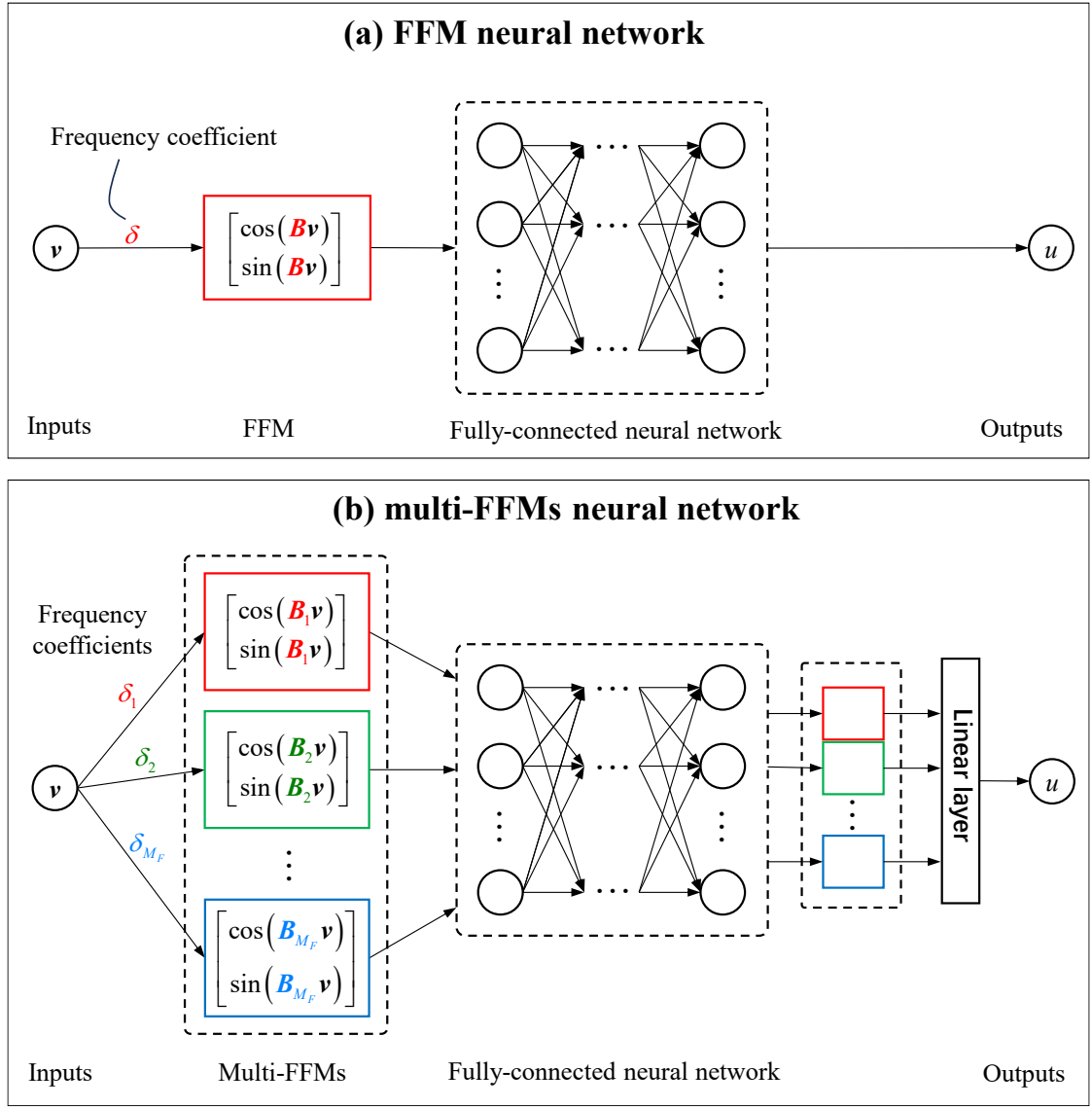

Figure 2 Schematic diagram of FFM neural network and multi-FFMs neural network: (a) FFM neural network; (b) multi-FFMs neural network

### 2.3. Multi-Fourier feature mappings (multi-FFMs)

To better deal with the problems whose solutions may contain different frequency components, it is reasonable to employ networks with multi-FFMs. As illustrated in Figure 2 (b), the input $v$ is first mapped by multi-FFMs initialized with different $\delta$ values, where $M_F$ denotes the number of FFMs. These transformed features are subsequently passed through a shared fully connected network and aggregated using a linear layer. Meanwhile, Li et al. (2024) proposed another architecture, where each FFM serves as the first hidden layer of a separate subnetwork. The outputs of all subnetworks are then concatenated with a linear layer.

While FFM as well as multi-FFMs is an efficient and powerful technique, it still suffers from certain inherent limitations. We must judiciously select both the number (e.g., $M_F$) and the scale (e.g., $\delta$ and $m_F$ in $\boldsymbol{B}$) of FFMs so that the frequency of the NTK eigenvectors aligns with that of the target function. To address this, Jin et al. (2024) introduced a trainable FFM ($\boldsymbol{B}$ is trainable) for Fourier feature networks and proposed a Fourier warm-start method for initializing $\boldsymbol{B}$. The trainable FFM may enhance the performance to some extent since $\boldsymbol{B}$ can be automatically tuned during the training. However, initialization of $\boldsymbol{B}$ demands prior knowledge of the frequency characteristics inherent in the target solutions. Such information is unavailable in most cases (especially for the forward problems), making the initialization empirical or based on "guessing".

## 3. General Fourier feature PIELM

### *3.1. Is Fourier feature PIELM (FF-PIELM) sufficiently effective?*

Concerning the applications of FFM and PINNs, it is natural to consider whether FFM can be concatenated with the ELM network to improve the performance of PIELM. A possible architecture, referred to as FF-ELM, is the concatenation of multi-FFMs and ELM network, as shown in Figure 3. There are double hidden layer in this FF-ELM architecture: (i) the first hidden layer is the multi-FFMs initialized with different $\delta$; and (ii) the second is same as the single-hidden-layer ELM architecture. This architecture is easy to follow since we simply replace the deep neural network in Figure 2 (b) by a ELM network.

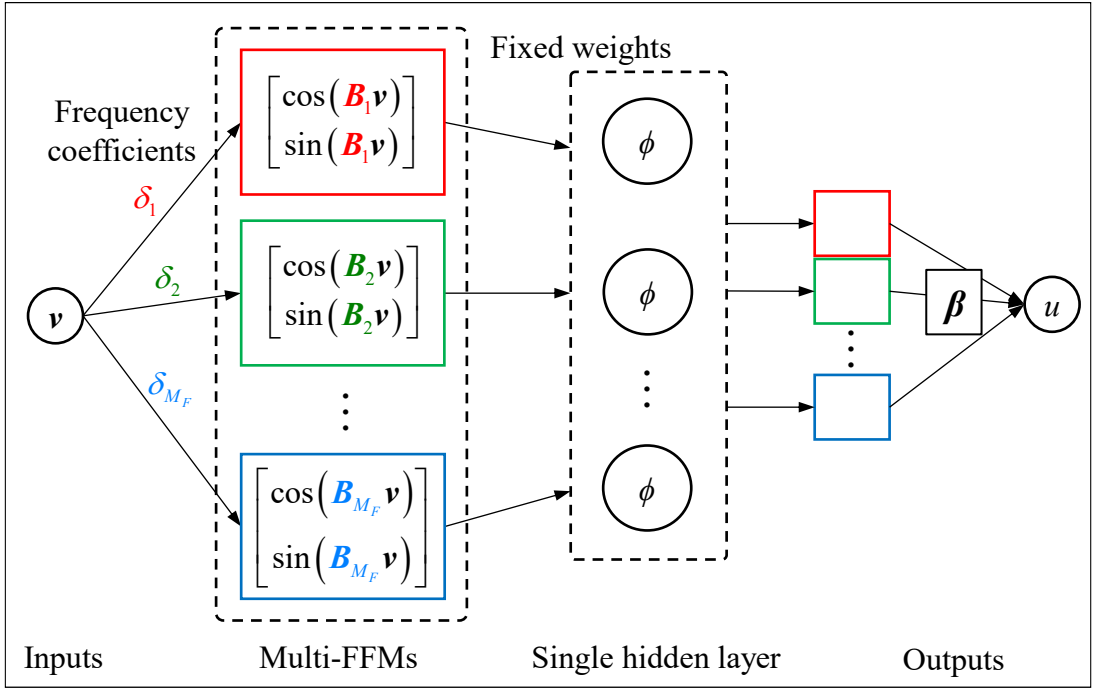

Figure 3 Schematic diagram of FFM-ELM architectures

After presenting a possible FF-ELM architecture, we will naturally ask: is this architecture sufficiently effective for constructing a powerful FF-PILEM? Unfortunately, the limitations outlined in Subsection 2.3 still persist in this FF-ELM architecture. The trainable FFM method cannot be applied to ELM network because the input weights are inherently untrainable. It is necessary to carefully tune a set of hyperparameters including $M_F$, $m_F$ and $\{\delta_i\}_{i=1}^{M_F}$ to match the frequencies of target solutions, but prior knowledge of these frequencies is not available in most cases. The challenge becomes even greater for solving variable-frequency PDEs, where an entire frequency range must be considered during initialization, rather than some specific frequencies.

### 3.2. GFF-ELM architecture for GFF-PIELM framework

To address the limitations of FF-PIELM, we need to further improve the network architecture and propose an effective initialization method. Before presenting our GFF-PIELM framework, let us first review the FFM function in Eq. (8). A variant of FFM (Wong et al. 2022) can be obtained by removing the sine function in Eq. (8) because the sine function can be transformed into a cosine function by a phase shift. Therefore, the FFM function can be rewritten as:

$$\gamma(v) = \cos(Bv + b) \qquad (9)$$

where $B \in \mathbb{R}^{m_F \times (d+1)}$ is initialized from the Gaussian distribution $\mathcal{N}(0, \delta^2)$, and $b$ is uniformly initialized within $[0, 2\pi]$. Compared with the conventional FFM, this variant is in fact more suitable for PIELM framework, as demonstrated below.

Let us substitute Eq. (9) into Eq. (4) by setting the FFM function as the activation function. Then we can get a Fourier-based ELM network in which the output of the $m$-th hidden neuron is $h_m(x,t) = \cos(w_m^T [x,t]^T + b_m)$. Interestingly, a neuron with Fourier-based activation function is now equivalent to a FFM with $m_F = 1$. This shows a possibility that $M$ neurons in the single-layer ELM network can be transformed into the multiple-FFMs ($M_F = M$). Hence, we introduce a set of frequency coefficients $\delta = [\delta_1, \delta_2, ..., \delta_M]$ and an $M \times M$ matrix $\Lambda = \text{diag}(\delta)$ into the hidden layer of ELM network, and then the latent solution will be approximated by:

$$u(x,t;\beta) = \sum_{m=1}^{M} \beta_m \cos(\delta_m w_m^T [x,t]^T + b_m) = \left[\cos(\Lambda W^T [x,t]^T + b)\right]^T \beta \qquad (10)$$

In Eq. (10) $W$ is initialized from the Gaussian distribution $\mathcal{N}(0,1)$, and $b$ is initialized uniformly within $[0, 2\pi]$. We refer to $\delta$ as the frequency coefficient vector, defined as a linearly spaced vector of $M$ points in the interval $[\delta_1, \delta_M]$, where $\delta_1$ and $\delta_M$ are user-defined hyperparameters. The proposed ELM architecture in Eq. (10) is referred to as GFF-ELM architecture, and illustrated in Figure 4.

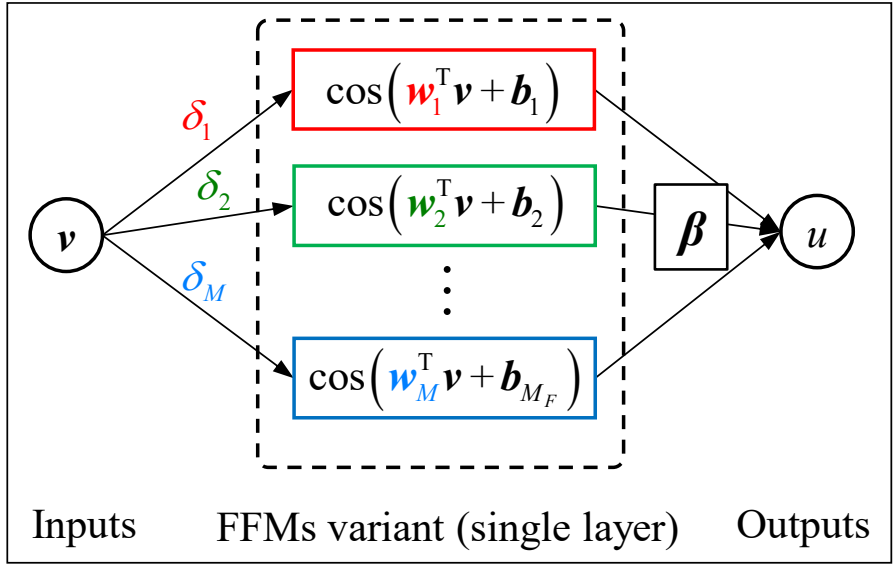

**Figure 4 Schematic diagram of general Fourier feature ELM**

In the GFF-ELM architecture each hidden neuron can be regarded as a variant of FFM with $\delta = \delta_m$ and $m_F =1$, and the single hidden layer is transformed into the multiple FFMs. In this way, the GFF-ELM architecture retains the simplification of ELM architecture while incorporating the properties and capabilities of FFM technique. Finally, the GFF-PIELM framework can be formed by replacing the ELM network in the vanilla PIELM framework with the novel GFF-ELM architecture. GFF-PIELM retains the advantages of GFF-ELM architecture, and two additional hyperparameters $\delta_1$ and $\delta_M$ are involved for initialization of the frequency matrix (e.g., $\mathbf{\Lambda W}^T$ in Eq. (10)). In the next subsection we will detail an innovative initialization method for selecting proper $\delta_1$ and $\delta_M$.

### *3.3. Initialization method for frequency-related hyperparameters*

The user-defined interval [$\delta_1$, $\delta_M$] specifies the frequency range that ELM prefers to learn, which should cover the frequency components of the target function. If the interval is too narrow, certain frequency components cannot be captured; if it is too wide, many hidden neurons become irrelevant and the useful neurons (with appropriate frequency coefficient $\delta$) are insufficient. As discussed in Subsection 2.3, the initialization of FFM is commonly based on experience or "guessing" owing to the lack of frequency information for the target PDE solution. To address this, we propose a new initialization method for GFF-PIELM as follows.

To illustrate the proposed initialization method for frequency coefficients, we use a simple one-dimensional (1D) Poisson's equation:

$$\frac{\partial^2 u(x)}{\partial x^2} = -9\pi^2 \sin(3\pi x) - 720\pi^2 \sin(60\pi x), \quad x \in [0,1] \tag{11}$$

$$u(0) = 0 \tag{12}$$

$$u(1) = 0 \tag{13}$$

The fabricated exact solution is

$$u(x) = \sin(3\pi x) + 0.2\sin(60\pi x) \tag{14}$$

The exact solution exhibits low-frequency behavior at the macro-scale and high-frequency oscillations at the micro-scale. The unknown solution $u(x)$ is represented by an ELM network with 200 hidden neurons. 400 collocation points sampled within (0, 1) and 2 boundary points are used for training.

The training performance is evaluated by mean squared error (MSE), given by

$$\text{MSE} = \frac{1}{N} \|\boldsymbol{H}\boldsymbol{\beta}^* - \boldsymbol{Y}\|_2^2 \tag{15}$$

where $N = N_C + N_B + N_I$ is the total number of training points. To quantify the prediction accuracy, the absolute error and relative $L_2$ error are defined respectively as

$$\text{Absolute Error} = |\text{exact} - \text{predicted}| \tag{16}$$

$$L_2 = \frac{\sqrt{\sum_1^N (\text{exact} - \text{predicted})^2}}{\sqrt{\sum_1^N (\text{exact})^2}} \tag{17}$$

When solving the one-dimensional Poisson's equation using our GFF-PIELM, we need to choose the interval $[\delta_1, \delta_M]$ for initialization. We first use a wide interval, e.g., $\delta_1 = 1$ and $\delta_M = 1000$, and the predicted solution is shown in Figure 5. As expected, GFF-PIELM fails to approximate the correct solution, as $\delta_1$ and $\delta_M$ may not be appropriate. In GFF-PIELM each hidden neuron is associated with a frequency coefficient $\delta$, and the output weights $\boldsymbol{\beta}$ reflect the degree of relevance between the neurons and the target function. A large $|\beta|$ value implies that the corresponding neuron contributes significantly to the solution, while a small or nearly zero $|\beta|$ value indicates irrelevance. Therefore, the interval $[\delta_1, \delta_M]$ can be adaptively refined according to the distribution of $\boldsymbol{\beta}$, as shown in Figure 5(c):

(i) When $\delta > 400$, most $\beta$ values vanish or are very close to zero, meaning that the neurons with $\delta > 400$ are ineffective.

(ii) When $\delta < 400$, some $\beta$ exhibit excessively large magnitudes, because the number of useful neurons is insufficient.

To remedy this, we increase the proportion of useful neurons (with $\delta < 400$) and remove useless ones (with $\delta > 400$) by adjusting the value of $\delta_M$ to 400. Figure 6 shows the adjusted result initialized with $\delta_1 = 1$ and $\delta_M = 400$. An excellent performance is achieved now with an MSE of 1.90e-17 and a relative $L_2$ error of 1.30e-12. The distribution of $\boldsymbol{\beta}$ is also well balanced without excessively large values (see Figure 6(c)). In practice, such hyperparameters can be successfully determined by one or two iterations, which is acceptable due to the high training efficiency of PIELM frameworks.

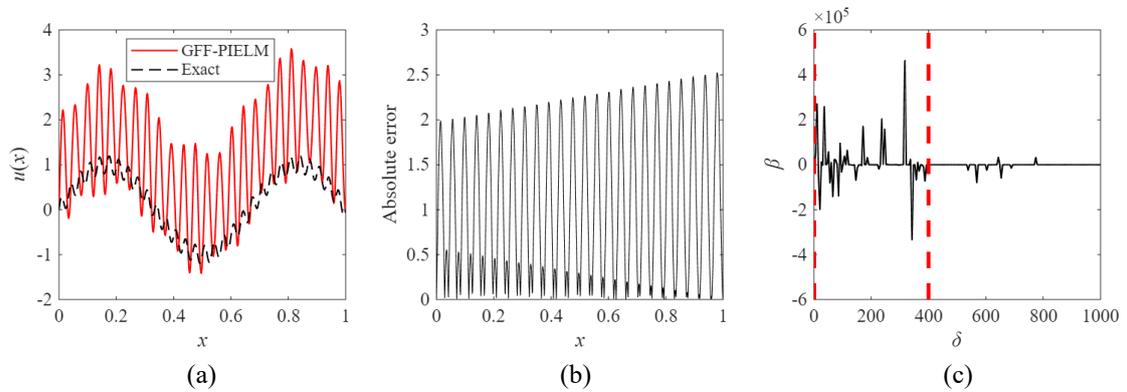

**Figure 5** GFF-PIELM for one-dimensional Poisson's equation with $\delta \in [1,1000]$: (a) GFF-PIELM versus exact solution; (b) Absolute error; (c) Distribution of $\boldsymbol{\beta}$ with $\delta$

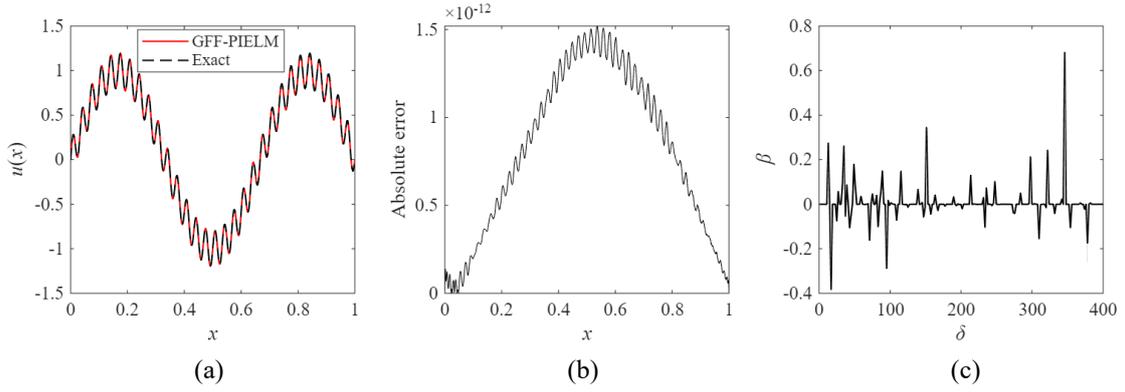

**Figure 6** GFF-PIELM for one-dimensional Poisson's equation with $\delta \in [1, 400]$: (b) GFF-PIELM versus exact solution; (b) Absolute error; (c) Distribution of $\boldsymbol{\beta}$ with $\delta$

## 4. Results and Performance

Ten typical examples in five case studies are conducted to show the performance of GFF-PIELM covering variable frequency, high frequency, complex-shape solution domains, parameter inversion and multi-scale. For comparison, the widely used vanilla PIELM with hyperbolic tangent (Tanh) activation function is also added, and optimal initialization of $L$ is determined using trial-and-error as shown in the Appendix. In all cases, the default setting for the two PIELM frameworks is: the number of hidden layer neurons is 5000; the training dataset includes 8000 random points inside the computational domain, 400 points on each boundary, and 400 points for each initial condition. Thus, the computational complexity is nearly identical across all cases, and the training time is approximately 9 seconds using MATLAB R2023b on a Lenovo X1 Carbon ThinkPad laptop with an Intel Core i7-1165G7 2.8GHz CPU and 16Gb of RAM memory. For easy reference, Table 1 lists the optimal initialization hyperparameters, MSEs and relative $L_2$ errors in the case studies for the vanilla PIELM and GFF-PIELM

**Table 1 Summary of optimal initialization hyperparameters, MSEs and relative $L_2$ errors for the vanilla PIELM and GFF-PIELM**

| Cases | Examples | Method | Initialization | MSE | $L_2$ error |
|---|---|---|---|---|---|
| Case 1: Wave equation | Linearly time-varying frequency | Vanilla PIELM | $L = 10$ | 0.16 | 0.55 |
| | | GFF-PIELM | $\delta_1 = 10, \delta_M = 100$ | 1.05e-09 | **3.41e-05** |
| | Periodically time-varying frequency | Vanilla PIELM | $L = 10$ | 3.62e-02 | 8.26e-02 |
| | | GFF-PIELM | $\delta_1 = 10, \delta_M = 150$ | 5.32e-06 | **1.62e-03** |
| Case 2: Wave equation | Fabricated solutions | Vanilla PIELM | $L = 10$ | 1.77e-04 | 0.49 |
| | | GFF-PIELM | $\delta_1 = 1, \delta_M = 100$ | 2.79e-11 | **1.09e-05** |
| | Series solution | Vanilla PIELM | $L = 10$ | 3.09e-05 | 0.12 |
| | | GFF-PIELM | $\delta_1 = 10, \delta_M = 140$ | 2.46e-08 | **2.44e-03** |
| Case 3: Helmholtz equation | Bat shape | Vanilla PIELM | $L = 10$ | 5.03e-03 | 0.18 |
| | | GFF-PIELM | $\delta_1 = 10, \delta_M = 110$ | 6.47e-13 | **3.04e-07** |
| | Monster shape | Vanilla PIELM | $L = 10$ | 1.16e-04 | 3.10e-03 |
| | | GFF-PIELM | $\delta_1 = 5, \delta_M = 60$ | 5.70e-09 | **2.35e-05** |
| Case 4: Klein Gordon equation | Forward problem | Vanilla PIELM | $L = 10$ | 1.20e-03 | 0.16 |
| | | GFF-PIELM | $\delta_1 = 20, \delta_M = 100$ | 1.56e-13 | **3.66e-07** |
| | Inverse problem | Vanilla PIELM | $L = 10$ | 2.47e-03 | 0.20 |
| | | GFF-PIELM | $\delta_1 = 20, \delta_M = 100$ | 1.09e-12 | **1.15e-06** |
| Case 5: Advection diffusion equations | One-dimensional | Vanilla PIELM | $L = 10$ | 1.67e-08 | 3.31e-05 |
| | | GFF-PIELM | $\delta_1 = 1, \delta_M = 100$ | 9.99e-19 | **3.71e-10** |
| | Two-dimensional | Vanilla PIELM | $L = 5$ | 2.70e-03 | 0.12 |
| | | GFF-PIELM | $\delta_1 = 1, \delta_M = 25$ | 5.12e-08 | **3.48e-04** |

## 4.1. Case 1: Variable-frequency wave equation

In this case we compare the performance of vanilla PIELM and GFF-PIELM using two examples for PDEs with variable frequency solutions. Let us consider the 1D wave equation taking the form:

$$\begin{aligned}
\frac{\partial^2 u(x,t)}{\partial t^2} - \frac{\partial^2 u(x,t)}{\partial x^2} &= f(x,t) \quad 0 \leq x \leq 1 \text{ and } 0 \leq t \leq 1 \\
u(0,t) &= g_1(t) \quad 0 \leq t \leq 1 \\
u(1,t) &= g_2(t) \quad 0 \leq t \leq 1 \\
u(x,0) &= u_0(x) \quad 0 \leq x \leq 1 \\
\frac{\partial u(x,0)}{\partial t} &= v_0(x) \quad 0 \leq x \leq 1
\end{aligned} \tag{18}$$

Two fabricated solutions are adopted: one exhibits the linearly time-varying frequency as

$$u(x,t) = \sin\left[(2\pi + 14\pi t)x\right]\cos(10\pi t) \tag{19}$$

and the other exhibits the periodically time-varying frequency as

$$u(x,t) = \sin\left[\pi\cos(4\pi t)x\right]\cos(4\pi t) \tag{20}$$

The source term $f(x, t)$ and the initial and boundary conditions are specified by the fabricated solutions.

Figure 7 and Figure 8 show the solutions and errors predicted by vanilla PIELM and GFF-PIELM for the two examples. In addition, the distribution of $\boldsymbol{\beta}$ with respect to $\delta$ is plotted to illustrate how the frequency-related hyperparameters ($\delta_1$ and $\delta_M$) are initialized in the GFF-PIELM framework. While the vanilla PIELM fails to accurately solve the PDEs, the GFF-PIELM achieves highly precise solutions for both examples with the $L_2$ error in the order of 1e-03~1e-05 (see Table 1). The accuracy is improved by a factor of more than 15000 and 50 for problems with linearly and periodically time-varying frequency, respectively. This demonstrates the capability of GFF-PIELM to tackle variable-frequency PDEs that remain challenging for existing PIML frameworks.

Another advantage that we would like to emphasize again is the straightforward determination of initialization hyperparameters. As shown in Figure 7 and Figure 8, $\delta_1$ and $\delta_M$ can be directly inferred from the distribution of $\boldsymbol{\beta}$, whereas selecting $L$ in vanilla PILEM requires the trial-and-error method as shown in the Appendix.

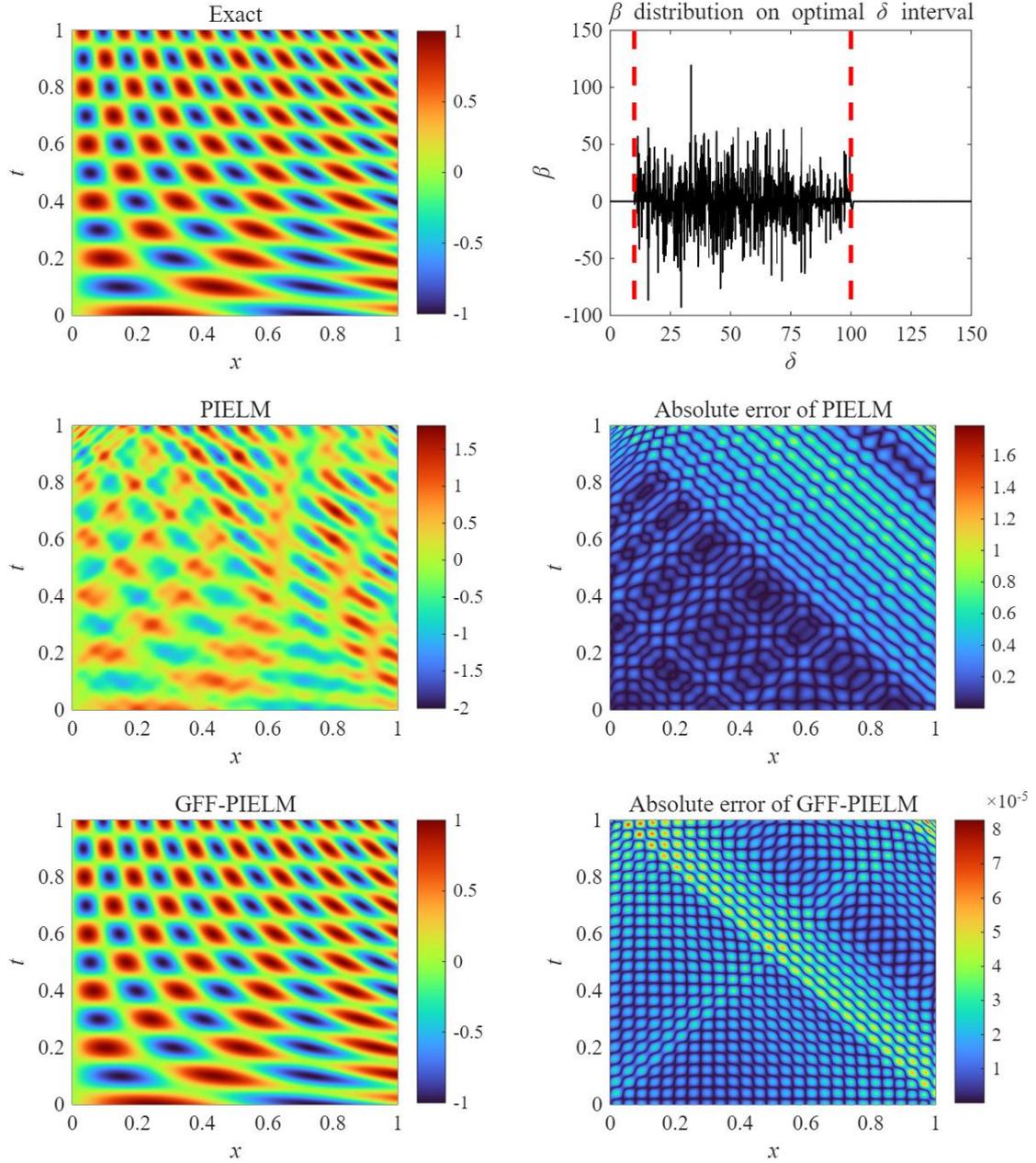

**Figure 7 Case 1: Comparison of PIELM and GFF-PIELM for wave equations with linearly time-varying frequency**

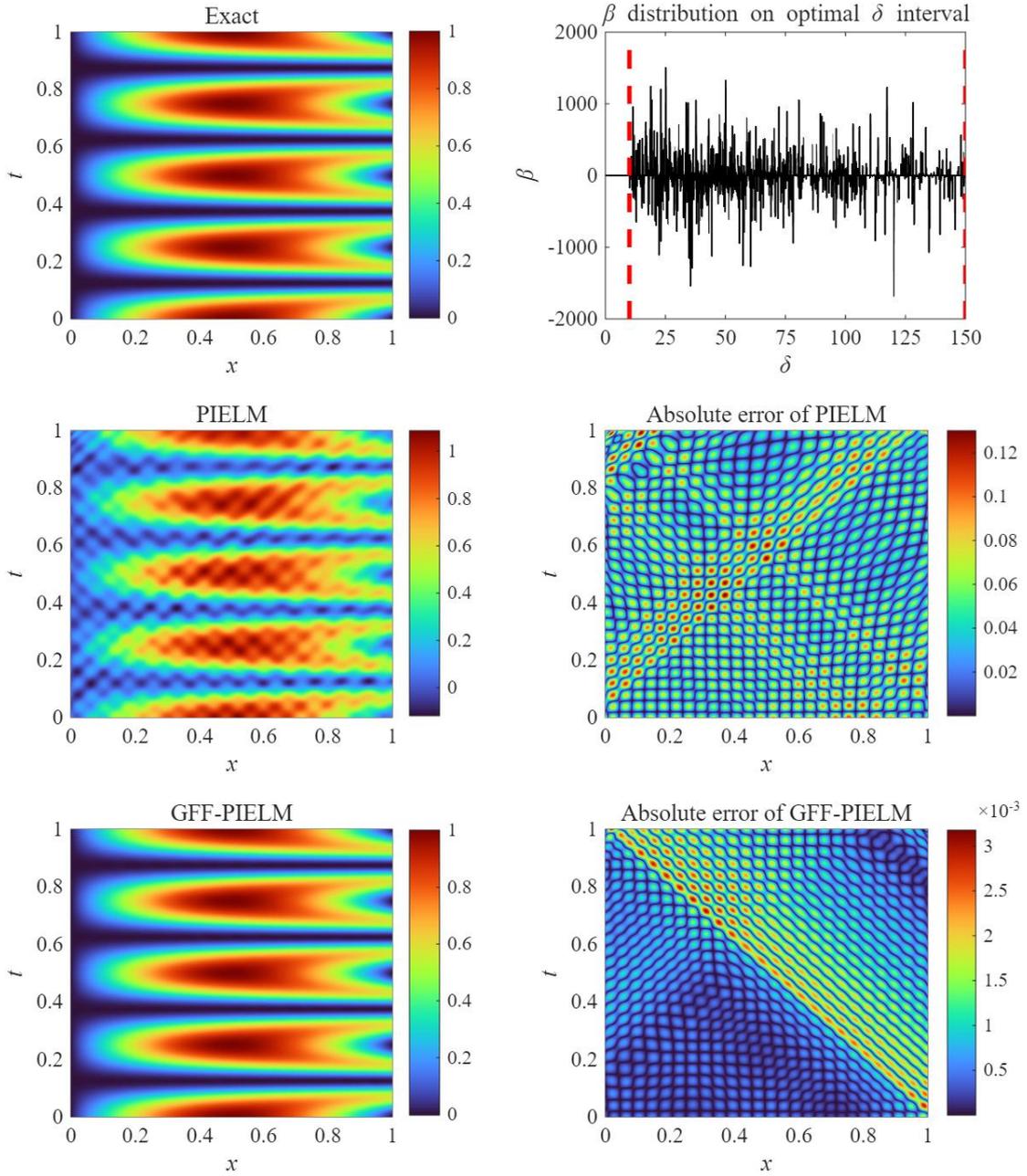

**Figure 8 Case 1: Comparison of PIELM and GFF-PIELM for wave equations with periodically time-varying frequency**

*4.2. Case 2: Multi-frequency wave equation*

In this case study we show the performance of GFF-PILEM in solving multi-frequency wave equations. The first example is a 1D wave equation taking the form:

$$\frac{\partial^2 u(x,t)}{\partial t^2} - 100\frac{\partial^2 u(x,t)}{\partial x^2} = 0 \quad 0 \le x \le 1 \text{ and } 0 \le t \le 1$$

$$u(0,t) = 0 \quad 0 \le t \le 1$$

$$u(1,t) = 0 \quad 0 \le t \le 1 \tag{21}$$

$$u(x,0) = \sin(\pi x) + \sin(2\pi x) \quad 0 \le x \le 1$$

$$\frac{\partial u(x,0)}{\partial t} = 0 \quad 0 \le x \le 1$$

The exact solution is

$$u(x,t) = \sin(\pi x)\cos(10\pi t) + \sin(2\pi x)\cos(20\pi t) \tag{22}$$

This example was previously investigated by Wang et al. (2021) using Fourier feature PINN. They reported that the PINN model cannot learn the correct solution, although a well-designed network architecture was employed. Therefore, they additionally employed an adaptive weights algorithm (Wang et al. 2022) to determine the loss weighting coefficients in the loss function, and then the accurate solution can be obtained with $L_2$=9.83e-04. By contrast, the PIELM-based GFF-PIELM approach does not require any treatment of loss weighting coefficients. Figure 9 shows the comparison of solutions and errors predicted by the vanilla PIELM and GFF-PIELM, and the other information is seen in Table 1. Again, the vanilla PIELM cannot learn the correct solution with acceptable accuracy, but the results of GFF-PIELM demonstrate excellent agreement between the predicted and the exact solution with $L_2$=1.09e-05. It is necessary to note that GFF-PIELM predict a similarly accurate solution to the work of Wang et al. (2021), whereas the PIELM-based approach significantly decreases training time from around 24minutes to 9s and does not require loss weighting coefficients.

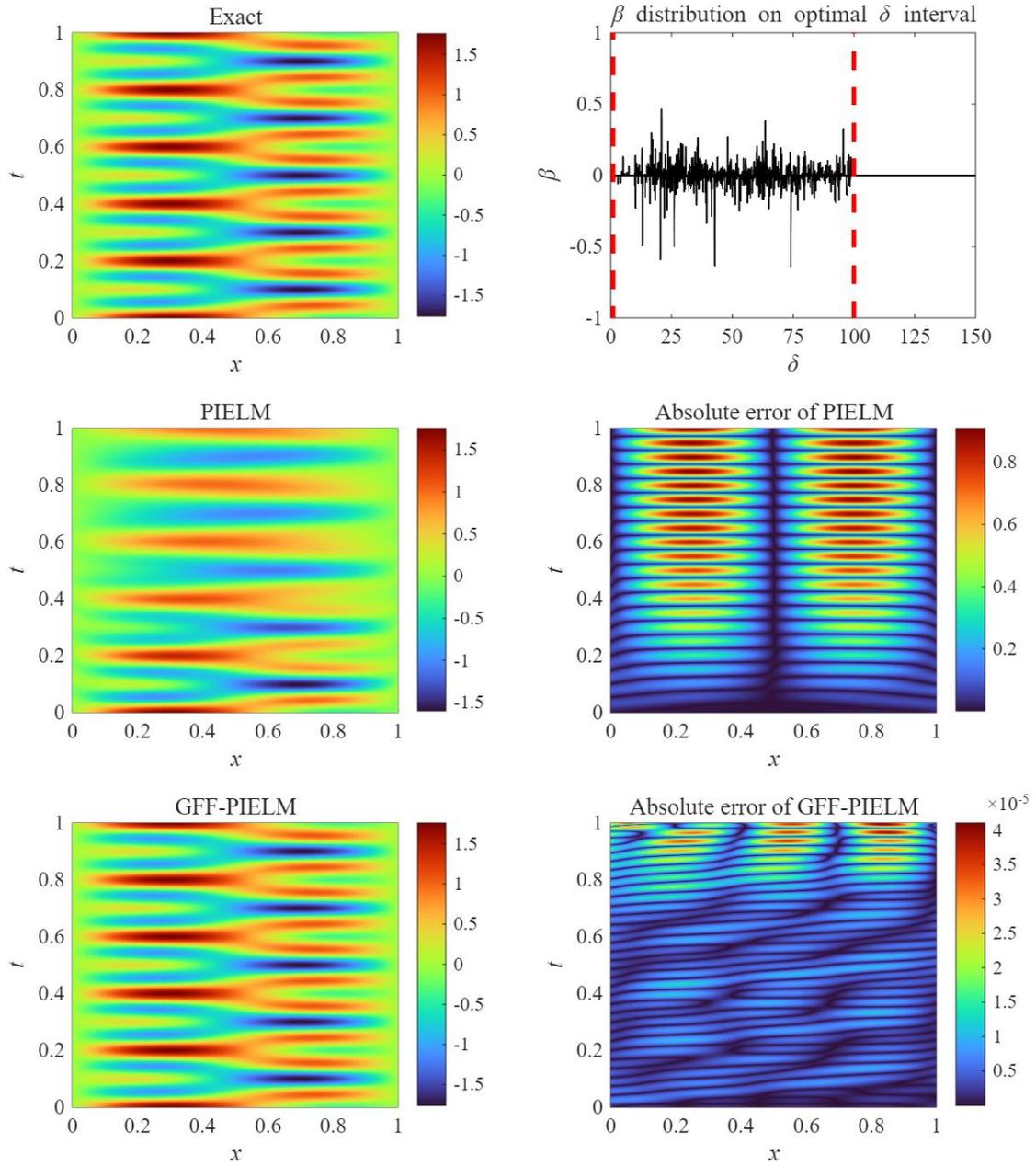

**Figure 9 Case 2: Comparison of PIELM and GFF-PIELM for multi-frequency wave equation**

Another challenging example is a 1D wave equation whose solution is expressed in series form. The equation is expressed as follows:

$$\frac{\partial^2 u(x,t)}{\partial t^2} - 49\frac{\partial^2 u(x,t)}{\partial x^2} = 0 \quad 0 \leq x \leq 1 \text{ and } 0 \leq t \leq 1$$
$$u(0,t) = 0 \quad 0 \leq t \leq 1$$
$$u(1,t) = 0 \quad 0 \leq t \leq 1 \tag{23}$$
$$u(x,0) = e^{\cos(\pi x)}\sin[\sin(\pi x)] \quad 0 \leq x \leq 1$$
$$\frac{\partial u(x,0)}{\partial t} = 0 \quad 0 \leq x \leq 1$$

The exact solution is given by

$$u(x,t) = \sum_{n=1}^{\infty} \frac{1}{n!}\cos(7n\pi t)\sin(n\pi x) \tag{24}$$

This series solution indicates that the exact solution comprises multiple frequency components. As the index $n$ increases, the coefficients progressively diminish, leading to the attenuation of the high-frequency terms. In this example, we retain the first 20 terms of the series solution ($n = 20$). Figure 10 shows the predicted solutions by vanilla PIELM and GFF-PIELM, and Table 1 lists the optimal initialization hyperparameters, MSE and $L_2$. We can see that the $L_2$ predicted by vanilla PIELM is just 0.12, while the GFF-PIELM gives a significantly more accurate prediction with $L_2$ error 2.44e-03.

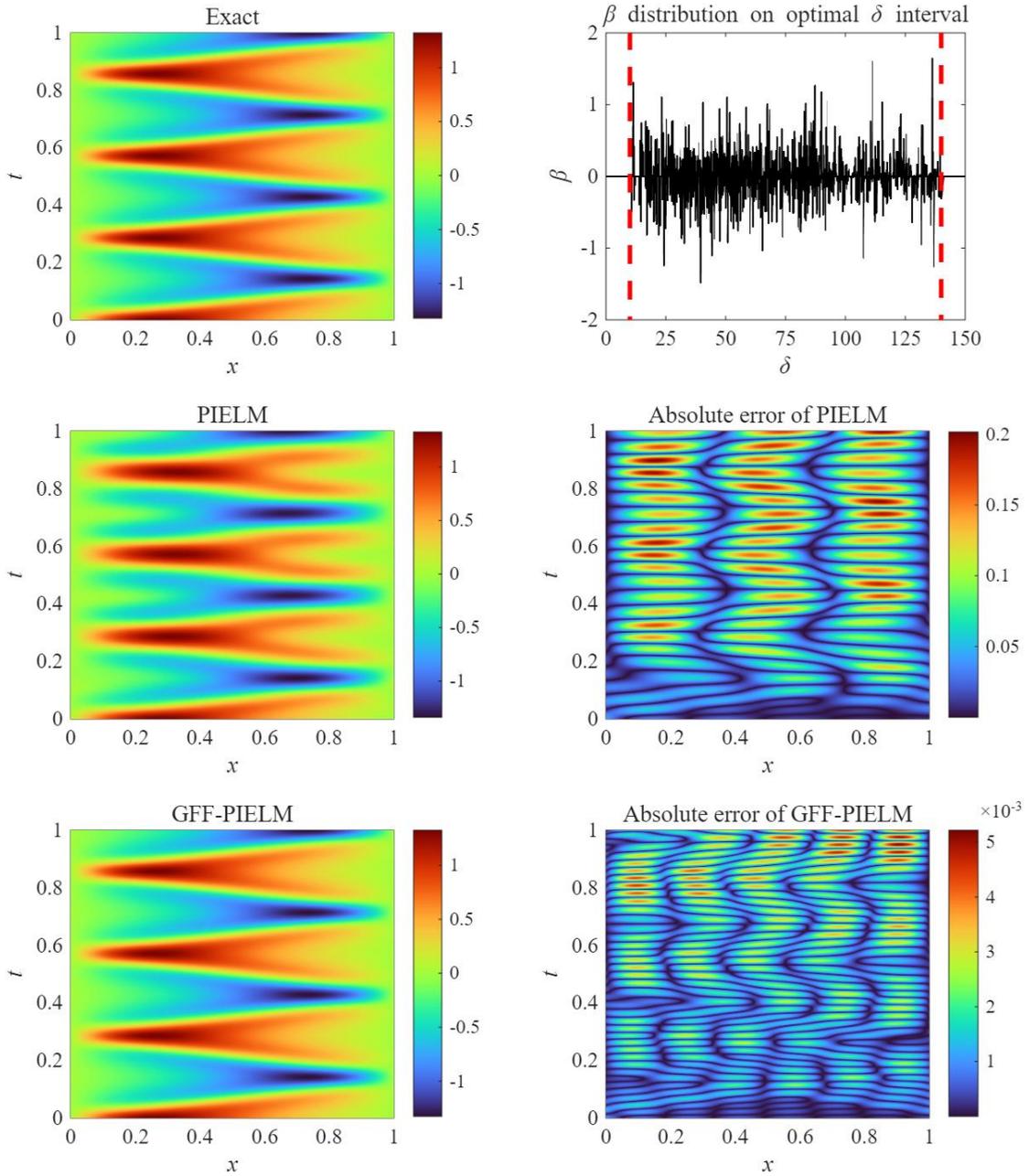

**Figure 10 Case 2: Comparison of PIELM and GFF-PIELM for wave equation with series solution**

### *4.3. Case 3: Helmholtz equation in complex solution domains*

In this case, the GFF-PIELM in solving PDEs in complex computational domains is highlighted using the Helmholtz equation. Let us consider the 2D Helmholtz equation taking the form:

$$\frac{\partial^2 u(x,y)}{\partial x^2} + \frac{\partial^2 u(x,y)}{\partial y^2} + u(x,y) = f(x,y) \quad (x,y) \in \Omega \tag{25}$$

Two examples are considered by giving two different fabricated solutions:

$$u(x,y) = \sin(25\pi x)\left[0.1\sin(8\pi y) + \tanh(8y)\right] \tag{26}$$

$$u(x,y) = \sin(2\pi x)\cos(4\pi x) + 0.5\sin(8\pi x)\cos(16\pi x) \tag{27}$$

In the first example, the fabricated solution in Eq. (26) exhibits sinusoidal behavior in the $x$ direction and a steep change along $y = 0$. In the second example, the fabricated solution in Eq. (27) shows multi-scale behavior in both $x$ and $y$ directions.

We select a bat-shaped solution domain for the first example (Eq. (26) is used as the solution of Eq. (24)) and a monster-shaped domain for the second example (Eq. (27) is adopted). The training data are sampled within these irregular domains and then fed into the PIELM frameworks. The source term $f(x, y)$ and the Dirichlet boundary conditions are specified by the fabricated solutions. Figure 11 and Figure 12 compare the predicted solutions obtained by vanilla PIELM and GFF-PIELM, and the training information and errors are summarized in Table 1. For the vanilla PIELM, the absolute error is on the order of 1e-01 and 1e-02 in two examples, respectively. The lower accuracy in the first example may be caused by the bat-shaped domain, the high-frequency behavior, and the steep change along $y = 0$, making the problem particularly challenging for vanilla PIELM. In terms of the GFF-PILEM, stable results are provided in both cases, and the solution accuracy reaches the order of 1e-7 and 1e-5, respectively. These results suggest that while PILEM can provide reasonable predictions (e.g., the trend is correct), incorporating the multi-FFMs is necessary for accurately solving high-frequency PDEs in complex domains.

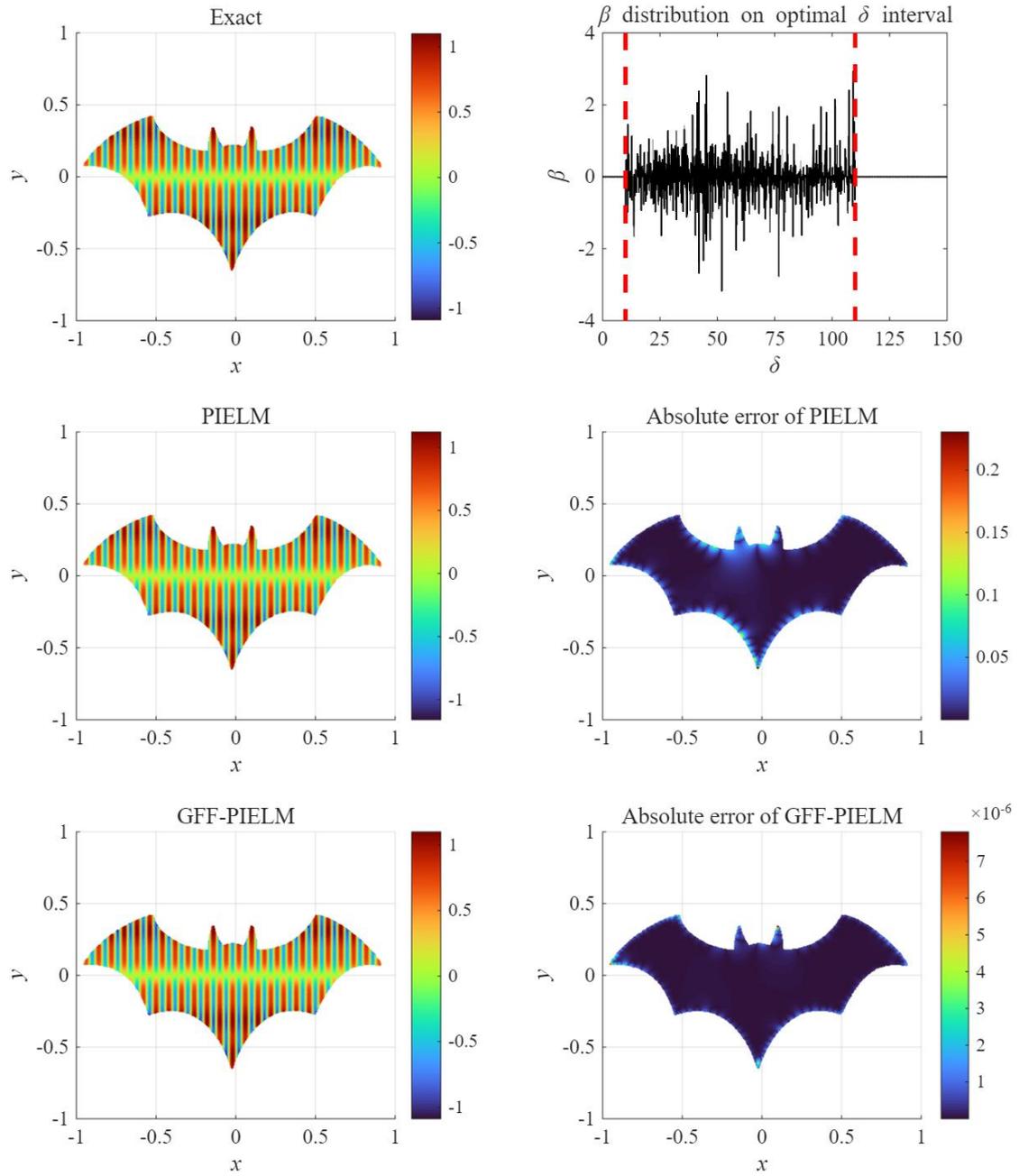

**Figure 11 Case 3: Comparison of PIELM and GFF-PIELM for Helmholtz equation in bat-shaped domain**

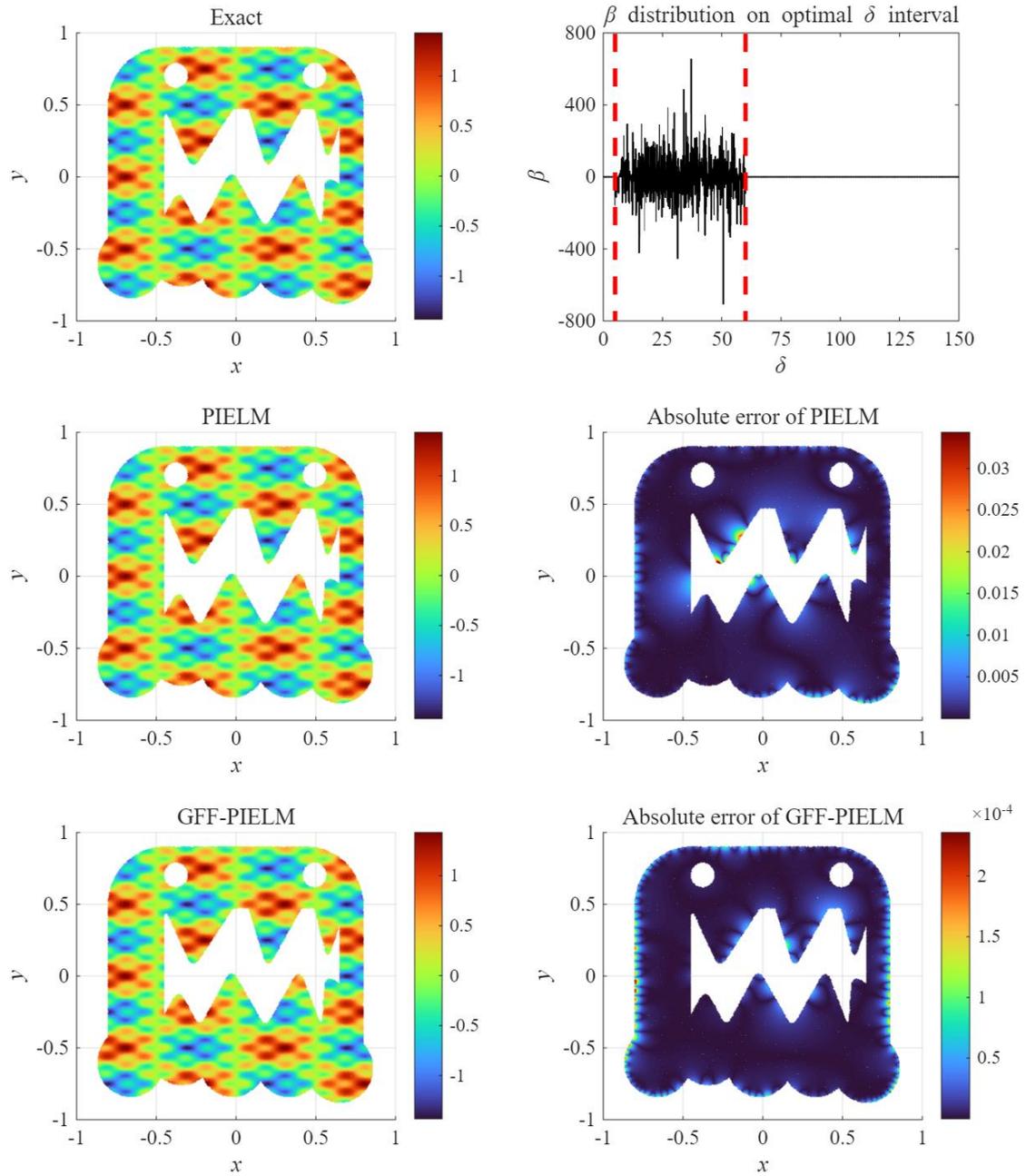

**Figure 12 Case 3: Comparison of PIELM and GFF-PIELM for Helmholtz equation in monster-shaped domain**

*4.4. Case 4: Klein Gordon equation: forward and inverse analyses*

In this case, we test GFF-PILEM in solving the Klein Gordon equation for forward and inverse analyses. For forward analysis, we consider the 1D linear Klein Gordon equation taking the form

$$\frac{\partial^2 u(x,t)}{\partial t^2} - \frac{\partial^2 u(x,t)}{\partial x^2} + u(x,t) = f(x,t;\alpha) \quad 0 \leq x \leq 1 \text{ and } 0 \leq t \leq 1$$

$$u(0,t) = g_1(t) \quad 0 \leq t \leq 1$$
$$u(1,t) = g_2(t) \quad 0 \leq t \leq 1 \tag{28}$$
$$u(x,0) = u_0(x) \quad 0 \leq x \leq 1$$
$$\frac{\partial u(x,0)}{\partial t} = v_0(x) \quad 0 \leq x \leq 1$$

The fabricated solution is given by

$$u(x,t) = x\sin(3\pi x)\cos(7\pi t) + t\sin(19\pi x)\cos(19\pi t) + \alpha x t \tag{29}$$

The source term $f(x, t; \alpha)$ and the initial and boundary conditions are determined from the fabricated solution with the parameter $\alpha=1$. For inverse analysis, $\alpha$ in the governing equation is treated as an unknown parameter and set as an additional output weight of PIELM frameworks. We assume that the initial and boundary conditions are known, and randomly generate 10 additional labelled points inside the computational domain using the solution Eq. (29) with the exact value $\alpha = 1$. The objective for inverse analysis is to simultaneously recover the solution for Eq. (28) as well as the parameter $\alpha$.

Figure 13, Figure 14 and Table 1 show the results for the forward and inverse Klein Gordon equation. The accuracy of GFF-PIELM in forward analysis is improved by a factor of more than 1e+06 compared to vanilla PIELM, with the $L_2$ error reaching the order of 1e-07. For inverse analysis, GFF-PIELM predicts $\alpha=1.00$, but vanilla PIELM provide $\alpha=2.17$ that is far from the true value. It is proven that as few as 10 additional labelled data are sufficient for GFF-PIELM to infer the unknown parameter $\alpha$, while maintaining high prediction accuracy for the solution.

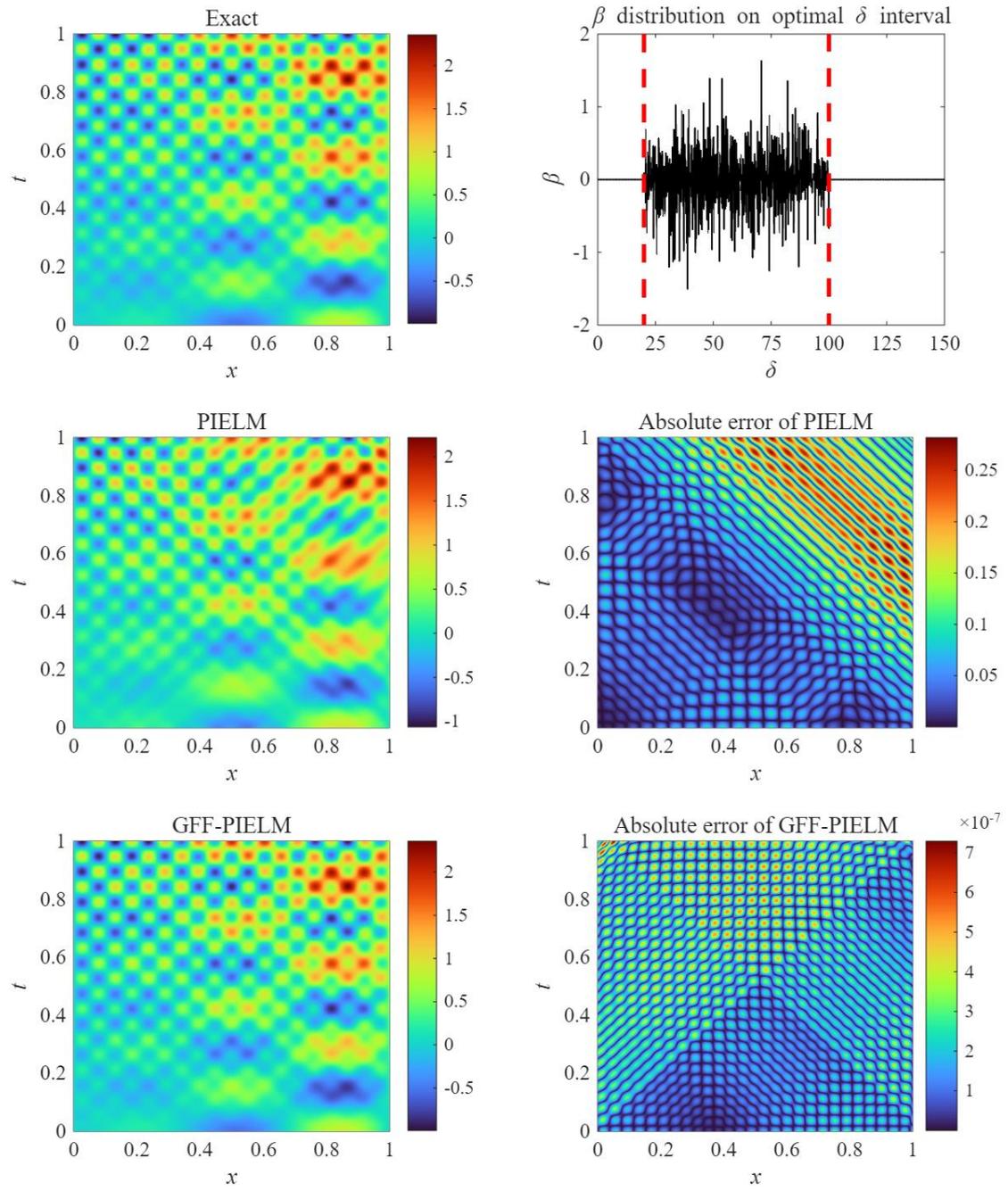

**Figure 13 Case 4: Comparison of PIELM and GFF-PIELM for Klein Gordon equation (Forward problem)**

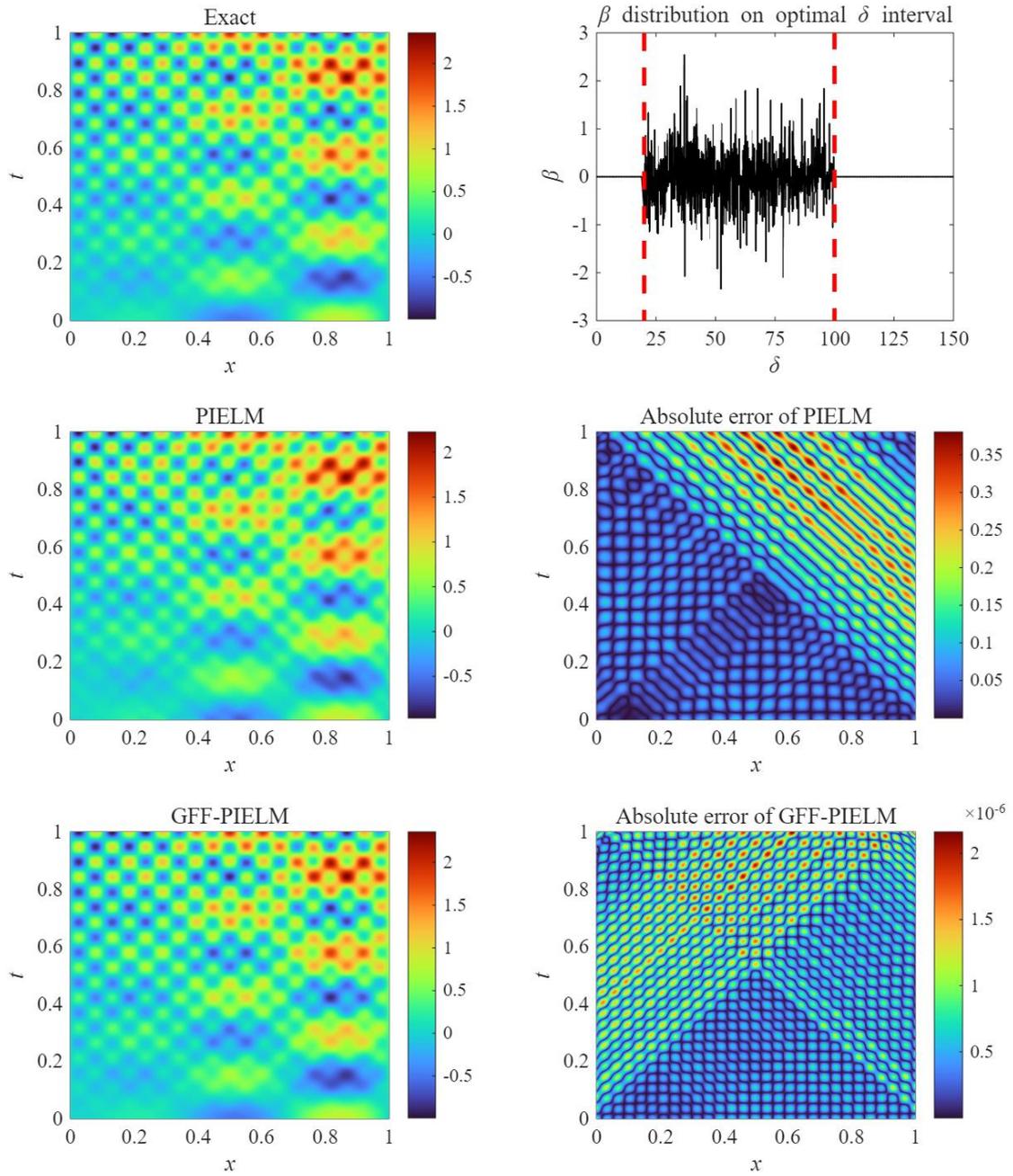

**Figure 14 Case 4: Comparison of PIELM and GFF-PIELM for Klein Gordon equation (Inverse problem)**

*4.5. Case 5: 1D and 2D Advection diffusion equations*

In the final case both 1D and two-dimensional (2D) advection diffusion equations are used to highlight the importance of GFF-PIELM. Let us first consider the 1D advection diffusion equation:

$$\frac{\partial u(x,t)}{\partial t} - 0.002\frac{\partial^2 u(x,t)}{\partial x^2} + 0.001\frac{\partial u(x,t)}{\partial x} = f(x,t) \quad 0 \leq x \leq 1 \text{ and } 0 \leq t \leq 1 \quad (30)$$

In the 1D example, the source term $f(x, t)$ and the initial and Dirichlet boundary conditions are specified by the following fabricated solution:

$$u(x,t) = e^{-0.5t}\left[\sin(\pi x) + 0.05\sin(25\pi x)\right] \quad (31)$$

The 2D advection diffusion equation we considered is as follows:

$$\frac{\partial u(x,y,t)}{\partial t} + 4\frac{\partial u(x,y,t)}{\partial x} + 4\frac{\partial u(x,y,t)}{\partial y} - \left[\frac{\partial^2 u(x,y,t)}{\partial x^2} + \frac{\partial^2 u(x,y,t)}{\partial y^2}\right] = f(x,y,t) \quad (x,y) \in \Omega \text{ and } 0 \leq t \leq 1 \quad (32)$$

The fabricated solution adopted for Eq. (32) is given by

$$u(x,t) = e^{-0.4t}\sin(4\pi x)\sin(8\pi y) \quad (33)$$

In the 2D example we select a Pacman-shaped solution domain $\Omega$. The source term $f(x, y, t)$ and the initial and Dirichlet boundary conditions are specified by the solution Eq. (33).

The capability of the two PIELM frameworks in solving Advection diffusion equations is compared in Figure 15, Figure 16, Figure 17 and Table 1. The results indicate that both vanilla PIELM and GFF-PIELM can predict accurate results for the 1D Advection diffusion equation, with the relative $L_2$ error on the order of 1e-05 and 1e-10, respectively. This implies that the accuracy is improved by more than five orders of magnitude through the incorporation of the general FFM. However, vanilla PIELM gives much larger absolute error for the 2D example, as shown in Figure 16. This is mainly because the hidden neurons and training points are insufficient for the vanilla PIELM to handle the 2D example under the default setting. On the contrary, GFF-PIELM shows its effectiveness in solving higher-dimensional PDEs with irregular computational domains.

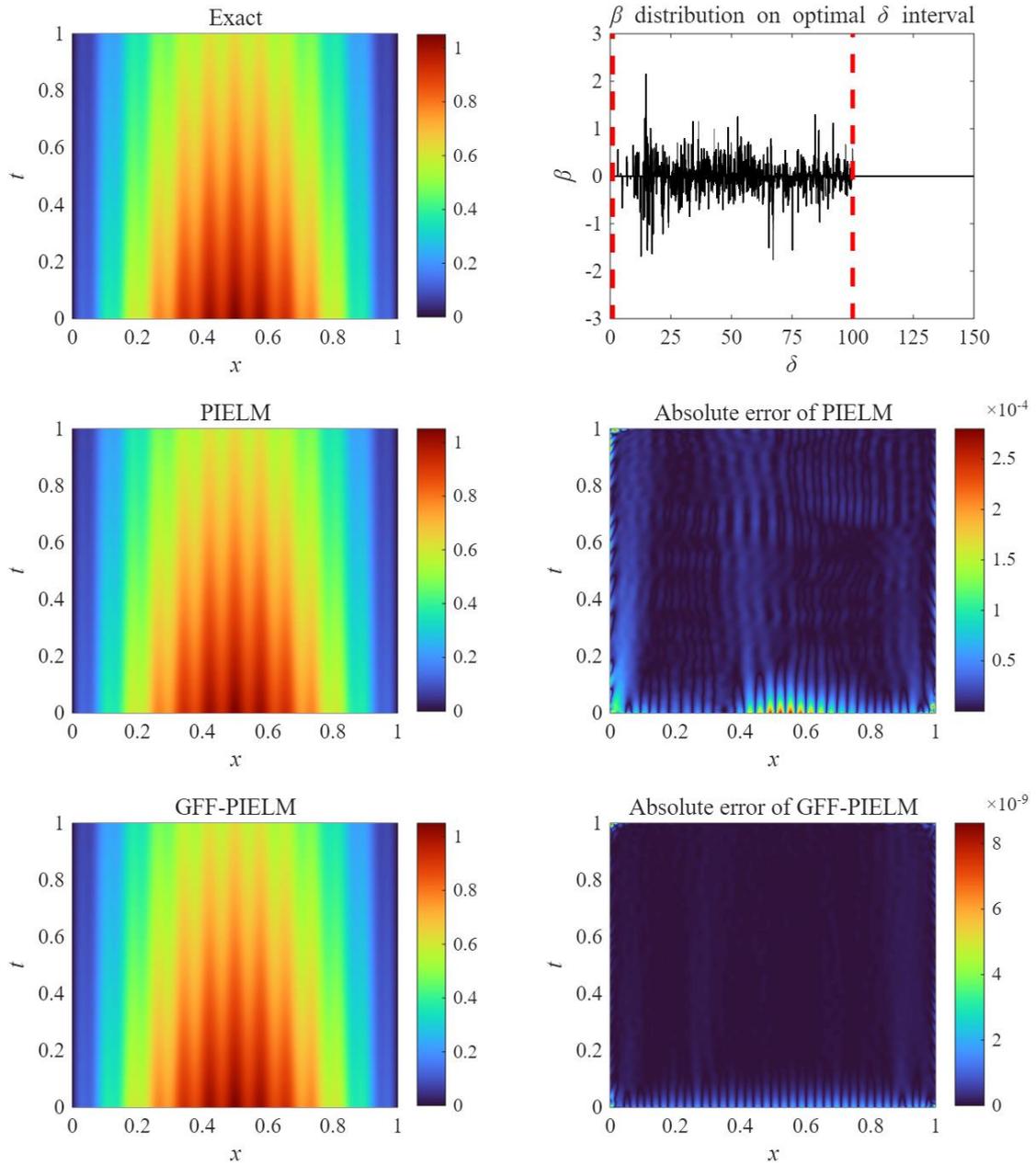

**Figure 15 Case 5: Comparison of PIELM and GFF-PIELM for 1D Advection diffusion equation**

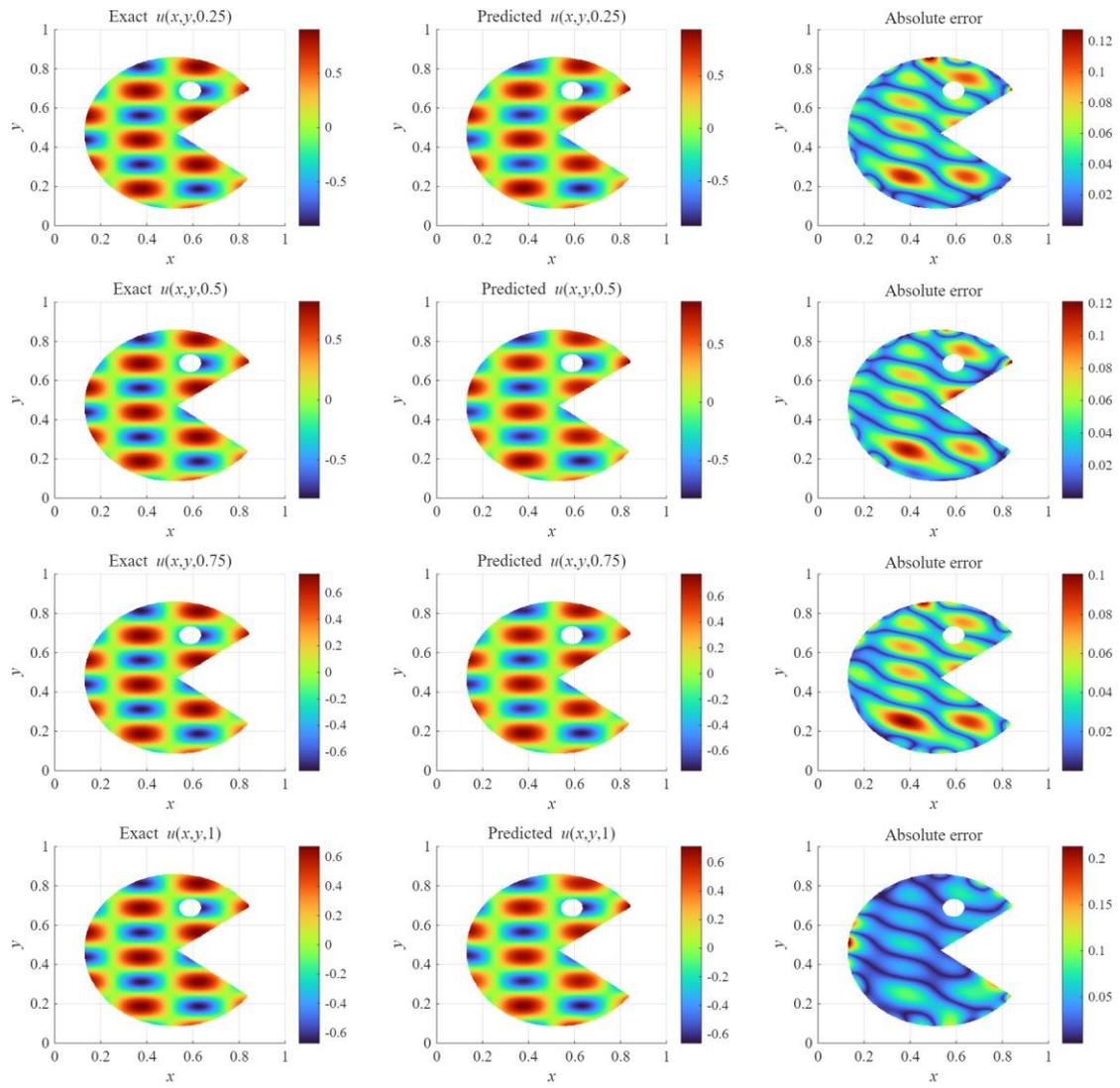

**Figure 16 Case 5: Vanilla PIELM for 2D Advection diffusion equation**

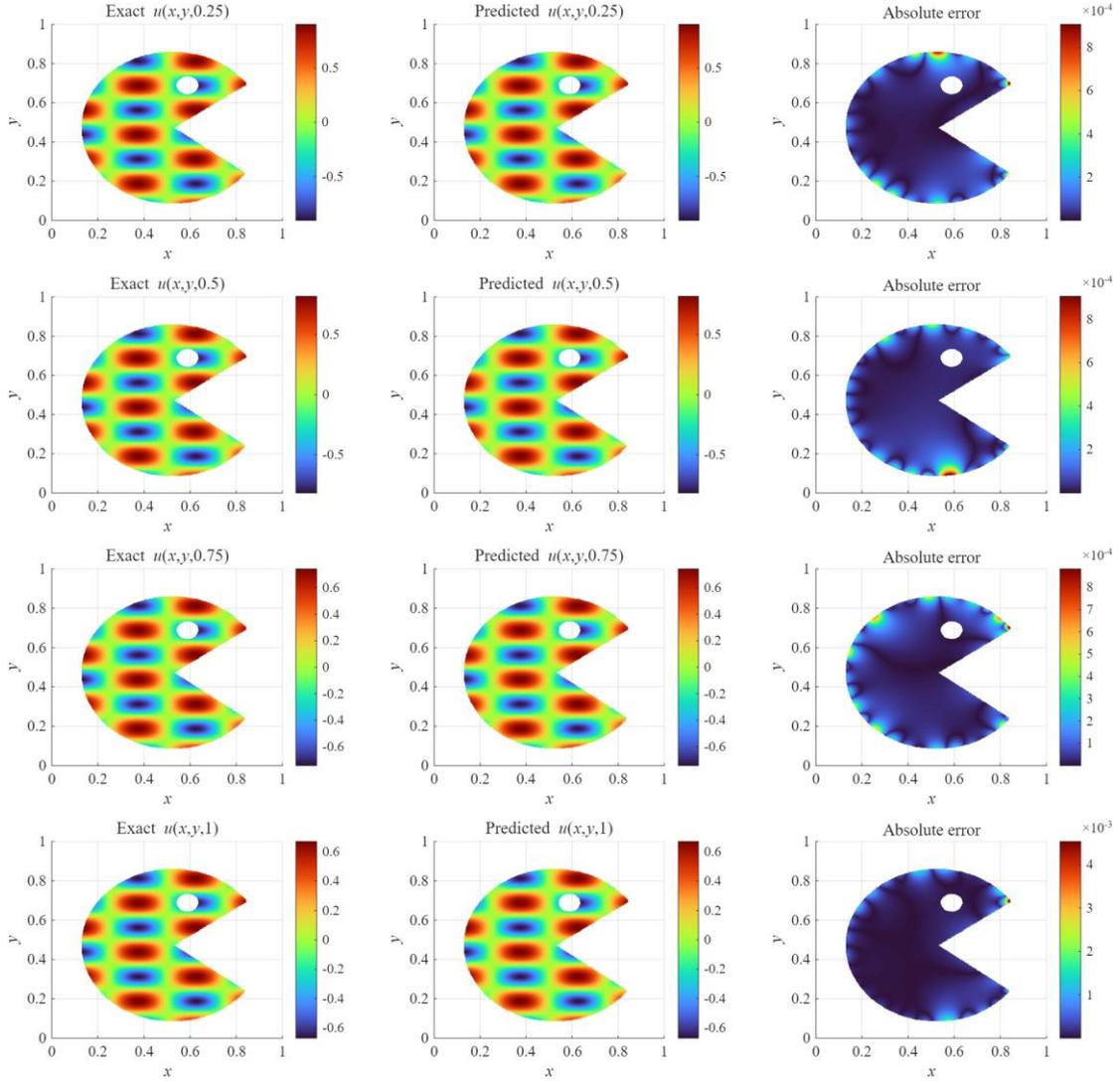

Figure 17 Case 5: GFF-PIELM for 2D Advection diffusion equation

## 5. Discussion

The superior performance of GFF-PIELM has been demonstrated by 10 examples in five case studies. Overall, the vanilla PIELM can provide reasonable predictions in certain examples, but it fails in most cases involving high frequency and variable frequency. This significant limitation is effectively alleviated by the proposed GFF-PIELM, which innovatively incorporates multi-FFMs into the PIELM framework. Compared to other popular approaches, GFF-PIELM have its own strength:

(i) Efficiency over PINNs. GFF-PILEM is based on the ELM network, and the training efficiency is significantly improved. In the above cases the training time required in GFF-PIELM is just several seconds, whereas conventional PINN typically requires approximately half an hour (one of the co-authors, Sifan Wang, shared the PINN codes at: https://github.com/PredictiveIntelligenceLab/MultiscalePINNs).

(ii) Accuracy over vanilla PIELM. Incorporating general FFMs greatly enhances the ability to capture the high-frequency and variable-frequency features of PDEs. The case studies in this paper also confirm the consistently higher accuracy of GFF-PIELM.

(iii) Flexibility over pseudo-spectral methods. GFF-PIELM can naturally handle irregular solution domains and inverse problems, as validated by several examples in this paper. This is mainly attributed to the fact that PIML is a mesh-free method and is highly convenient to integrate physical laws with data.

(iv) Simplicity of architecture. The single-layer ELM network combined with a variation of FFM ensures that GFF-PIELM remains simple and efficient. The initialization of frequency-related hyperparameters can be easily determined by inspecting the distribution of output weights, rather than by the trial-and-error tuning of hyperparameters..

Despite these advantages, GFF-PIELM is not without limitations. The inherent limitation of PIELM in solving PDEs with sharp gradients may also be applied to GFF-PIELM. In PINNs, such issues can sometimes be tackled by significantly increasing the number of training points to smooth the gradients, but this strategy is impractical for PIELM due to the lack of batch training. Recently, time-stepping methods have been incorporated into PIELM to smooth PDE gradients across successive batches, showing promise in handling high-frequency temporal behaviors (Yang et al. 2025). However, these methods have been applied only to parabolic PDEs, rather than hyperbolic ones (e.g., wave equations) where high-frequency and variable-frequency effects are often more critical in practice. Also, the time-stepping method for inverse analysis, such as inferring certain parameters, may also face difficulties.

Another major limitation is the relatively large number of hidden neurons required for acceptable performance. Although the scale of both the architecture and dataset in GFF-PIELM is substantially smaller than that in PINN frameworks, the computational burden grows significantly when addressing nonlinear PDEs. This is because for PIELM relies on iterative least-squares solvers for nonlinear problems, meaning the computational load is incurred at every iteration. Moreover, iterative least-squares approaches require initialization of output weights, and handling a large number of hidden neurons makes this initialization increasingly difficult. To address these challenges, integrating GFF-PIELM with advanced strategies such as time-stepping schemes, domain decomposition and curriculum Learning may be helpful (Dong and Li 2021; Dwivedi et al. 2025; Yang et al. 2025). These approaches can effectively reduce the scale of the network and dataset and constitute a promising direction for future research.

## 6. Conclusions

In this paper we propose a novel PIML framework, termed GFF-PIELM framework, for solving high-frequency and variable-frequency PDEs. By integrating a variant of multi-FFMs into the PIELM framework, GFF-PIELM effectively combines the high accuracy and training efficiency of PIELM with the capability of FFMs to capture high-frequency components. We also introduce a dedicated initialization strategy for frequency-related hyperparameters from the distribution of ELM output weights. The novel initialization strategy is able to determine the frequency range of the target solution in a straightforward way, thus avoiding the conventional trial-and-error procedure. Ten numerical experiments demonstrate that GFF-PIELM consistently outperforms vanilla PIELM, achieving improvements in accuracy while maintaining extremely low computational cost. The advantages of GFF-PIELM are highlighted as follows: (i) superior training efficiency compared to PINNs due to the use of ELM networks; (ii) enhanced ability to capture high-frequency and variable-frequency features relative to vanilla PIELM; (iii) capability to handle irregular solution domains and inverse problems compared to pseudo-spectral methods; and (iv) a simple architecture with easily determined hyperparameters. Overall, GFF-PIELM provides an accurate and efficient approach for solving challenging PDEs with high-frequency and variable-frequency behavior, showing significant potential for applications in computational science, engineering, and data-driven physical modeling.

## Acknowledgement

We acknowledge the funding support from the Natural Science Foundation of Shandong Province (Grant No. ZR2022QA046 and ZR2024LZN002).

## Data Available Statement

Data are available from the corresponding author upon reasonable request.

## Conflict of Interests

The authors declare that there is no known conflict of interest

## Declaration of generative AI use

We used large language models (LLMs) to help refine the writing and presentation of this paper (e.g., clarifying explanations, improving readability, and rephrasing drafts). All technical content—including problem formulation, method design, theoretical results, and experiments—was conceived, implemented, and validated by the authors.

## Appendix  Trial-and-error hyperparameter initialization in vanilla PIELM

In the appendix we show the initialization method for vanilla PIELM using the 1D

Poisson's equation in Subsection 3.3 as the example. The vanilla PIELM is equipped with hyperbolic tangent (Tanh) activation function, which is the most commonly used aperiodic activation. The input layer weights (**W** and **b**) are uniformly initialized within the interval [-$L$, $L$]. The performances with different initialization settings are shown in Table 2 and Figure 18, and the optimal performance is achieved at $L$=40. The numerical results indicate that the initialization can largely affect the performance of PIELM. Therefore, we need to tune $L$ by the trial-and-error method even though it is time-consuming.

Table 2  PIELM performance with different initialization settings

| Initialization setting | $L$ = 1 | $L$ = 20 | $L$ = 40 | $L$ = 60 |
| --- | --- | --- | --- | --- |
| MSE | 2.48e7 | 1.06e7 | 5.55e5 | 8.43e6 |
| $L_2$ error | 0.20 | 0.13 | 2.20e-2 | 0.24 |

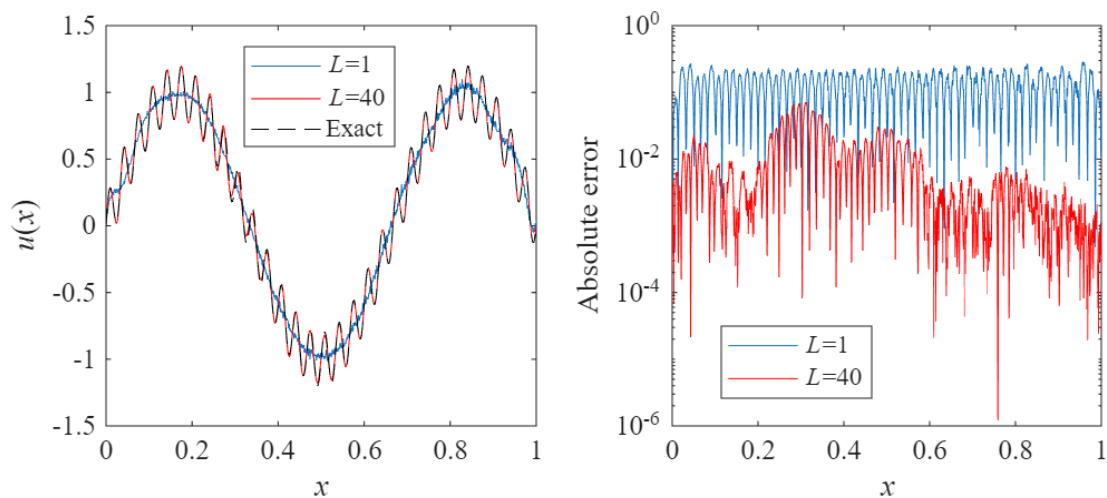

Figure 18 Exact solution and absolute error predicted by the vanilla PIELM for Poisson's equation